\documentclass{article}

\usepackage{arxiv}

\usepackage[utf8]{inputenc} 
\usepackage[T1]{fontenc}    
\usepackage{hyperref}       
\usepackage{url}            
\usepackage{booktabs}       
\usepackage{amsfonts}       
\usepackage{nicefrac}       
\usepackage{microtype}      
\usepackage{lipsum}
\usepackage{fancyhdr}       
\usepackage{graphicx}       
\graphicspath{{media/}}     
\usepackage{amsthm}
\usepackage{amsmath}       
\usepackage{amssymb}        
\usepackage{bm}  
\usepackage{float}
\usepackage{subcaption}
\newtheorem{theorem}{Theorem}
\usepackage{algorithm}
\usepackage{algpseudocode}
\usepackage{xcolor} 
\title{Conditional Prediction ROC Bands for \\ Graph Classification}

\author{
  Yujia Wu \\
  Center for Data Science \\
  New York University \\
  \texttt{yw7702@nyu.edu} \\
   \And
  Bo Yang \\
  Department of Biostatistics \\
  University of Michigan - Ann Arbor \\
  \texttt{ybb@umich.edu} \\
   \And
  Elynn Chen \\
  Stern School of Business \\
  New York University \\
  \texttt{elynn.chen@stern.nyu.edu} \\
   \And
  Yuzhou Chen \\
  Department of Statistics \\
  University of California, Riverside \\
  \texttt{yuzhou.chen@ucr.edu} \\
  \And
  Zheshi Zheng \\
  Department of Biostatistics \\
  University of Michigan - Ann Arbor \\
  \texttt{zszheng@umich.edu} \\
}

\begin{document}
\maketitle

\begin{abstract}
Graph classification in medical imaging and drug discovery requires accuracy and robust uncertainty quantification. To address this need, we introduce Conditional Prediction ROC (CP-ROC) bands, offering uncertainty quantification for ROC curves and robustness to distributional shifts in test data. Although developed for Tensorized Graph Neural Networks (TGNNs), CP-ROC is adaptable to general Graph Neural Networks (GNNs) and other machine learning models. We establish statistically guaranteed coverage for CP-ROC under {\it a local exchangeability condition}. This addresses uncertainty challenges for ROC curves under non-iid setting, ensuring reliability when test graph distributions differ from training data. Empirically, to establish local exchangeability for TGNNs, we introduce a data-driven approach to construct local calibration sets for graphs. Comprehensive evaluations show that CP-ROC significantly improves prediction reliability across diverse tasks. This method enhances uncertainty quantification efficiency and reliability for ROC curves, proving valuable for real-world applications with non-iid objects.
The code for this paper is publicly available at \url{https://github.com/CS-SAIL/UQ-GNN-ROC.git}.
\end{abstract}


\section{Introduction}
Graph classification is essential in fields like medical image analysis and drug discovery, where high accuracy and robust uncertainty quantification (UQ) are critical. Graph Neural Networks and Tensorized Graph Neural Networks excel in label prediction accuracy by leveraging complex graph and tensor structures, enhancing statistical performance and scalability~\cite{xia2021graph, zhou2020graph, wen2024tensor}. In addition, recent advancements in UQ for GNN include conformal methods for robust prediction sets~\cite{zargarbashi2023conformal, huang2023uncertainty}, specialized models for travel demand and traffic risk assessment~\cite{zhuang2022uncertainty, gao2022spatiotemporal}, and Bayesian frameworks for quantifying aleatoric and epistemic uncertainties~\cite{munikoti2023general}. These advancements enhance the accuracy and uncertainty quantification of graph label prediction. 

However, significant gaps remain. Firstly, simple UQ at the label level is insufficient for a comprehensive assessment of classifier reliability. The ROC curve, which provides a detailed illustration of classifier performance across thresholds, is especially useful in critical applications with imbalanced data and varying costs of false positives and negatives. Thus, the UQ of the ROC curve is needed for advanced graph classification tools like TGNNs. Secondly, most UQ for GNN does not address covariate shifts in graph features, which is common in large datasets. Existing methods for non-exchangeable cases, which adjust conformal prediction intervals for estimated covariate shifts~\cite{tibshirani2019conformal, gibbs2021adaptive}, are not directly applicable to GNNs where graph covariate shifts are hard to estimate and are not suitable for quantifying uncertainty of specific graph types.

To address these gaps, we propose Conditional Prediction ROC (CP-ROC) bands, which create prediction bands for ROC curves that are robust to distributional shifts in graph data. Developed for TGNNs, this method can be adapted to GNNs and other neural networks and machine learning models by modifying the similarity measure based on embedded features. Despite the challenge of constructing these prediction bands due to the complex structure of tensorized graph data, we design an efficient, data-driven method for calculating similarities between different tensorized graphs.

\noindent
\textbf{Our Contributions.} We make three key advances in UQ for graph classification:

(1) \underline{\textit{Conditional Prediction ROC (CP-ROC) Bands}}: We propose a novel method providing conditional confidence bands for ROC curves, the first of its kind for graph classification.

(2) \underline{\textit{Robust Coverage under Graph Covariate Shifts}}: We offer statistically guaranteed conditional confidence bands robust to graph covariate shifts. This approach, requiring a local calibration set of similar graph data, significantly enhances model adaptability under a ``local exchangeable'' condition.

(3) \underline{\textit{Empirical Method and Evaluation}}: We propose a data-driven approach to construct local calibration sets for TGNNs that ensure local exchangeability. Extensive evaluations on 6 datasets demonstrate the efficacy and robustness of our method.

\section{Related Work}
\textbf{Conditional Conformal Prediction for Deep Learning.} Conformal Prediction (CP) ~\cite{vovk2005algorithmic} generates prediction sets that include the true outcome with a specified probability $1-\alpha\in(0,1)$. 
CP's distribution-free nature makes it versatile in providing robust uncertainty estimates across various fields, including computer vision~\cite{angelopoulos2020uncertainty,bates2021distribution,angelopoulos2022uncertainty}, causal inference~\cite{lei2021conformal,jin2023sensitivity,yin2024conformal}, time series forecasting~\cite{gibbs2021adaptive,zaffran2022adaptive}, and drug discovery~\cite{jin2023selection}.

Recent work has applied conformal prediction to GNNs for node classification:\cite{zargarbashi2023conformal} propose a method using node-wise conformity scores, and \cite{huang2023uncertainty} introduce Conformalized Graph Neural Networks (CF-GNN) for rigorous uncertainty estimates. Our work differs by focusing on whole-graph classification using GNNs for representation learning. However, our proposed conditional prediction ROC bands can potentially integrate with these node classification methods.

``Conditional coverage'' in Conformal Prediction (CP) aims for coverage at specific covariate values, which is challenging without additional assumptions~\cite{lei2014distribution, vovk2012conditional, barber2021limits}. Approaches for approximate conditional coverage include improved score functions~\cite{romano2019conformalized, romano2020classification, angelopoulos2022image}, adapting to covariate shifts~\cite{romano2020malice, gibbs2023conformal}, and using similar calibration data~\cite{ding2023class, guan2023localized}. However, these methods face limitations with GNNs due to difficulty estimating graph covariate shifts, unsuitability for certain graph types. Thus, a new method for conditional UQ for the ROC is needed.

\textbf{UQ for ROC.} Confidence bands for the ROC curve can be constructed using either point-wisely \cite{schafer1994efficient,hilgers1991distribution} or globally \cite{jensen2000regional,campbell1994advances}.
We focus on point-wise confidence band in this work because the global bands are often too wide to be useful. Most existing work on ROC confidence bands is designed for diagnostic testing settings \cite{nakas2023roc}, which do not apply to our problem. Additionally, popular bootstrap-based UQ methods \cite{adler2009bootstrap} tend to underestimate uncertainty. Thus, we generalize the conformal-based method from \cite{zheng2024roc}, ensuring proper coverage of the oracle ROC curve.

\textbf{Graph Neural Networks.} GNNs are crucial for learning representations from graph-structured data, excelling in various applications~\cite{kipf2016semi,velickovic2017graph}. Key developments include message-passing mechanisms~\cite{gilmer2017neural} and attention-based models~\cite{velickovic2018graph}, addressing issues like over-smoothing and scalability~\cite{li2018deeper}. Despite their effectiveness, traditional GNNs lack robust uncertainty quantification, which is critical for applications requiring reliable confidence intervals. While Bayesian approaches and ensemble methods have been explored to address this, they often lack theoretical guarantees and increase computational complexity~\cite{zhang2019bayesian}.

\textbf{Tensor Learning.} Tensors, which represent multidimensional data, have gained prominence across various scientific disciplines, including neuroimaging \cite{zhou2013tensor}, economics \cite{chen2020constrained,liu2022identification}, international trade \cite{chen2022modeling}, recommendation systems \cite{bi2018multilayer}, multivariate spatial-temporal data analysis \cite{chen2020modeling}, and biomedical applications \cite{chen2024semi,chen2024distributed}. Empirical research has demonstrated the effectiveness of tensor-based classification methods in these fields, leading to the development of techniques such as support tensor machines \cite{hao2013linear, guoxian2016}, tensor discriminant analysis \cite{chen2024high}, tensor logistic regression \cite{wimalawarne2016theoretical}, and tensor neural networks \cite{kossaifi2020tensor,wen2024tensor}.

However, existing methods have two main limitations: they do not address graph-structured data and lack uncertainty quantification. Our research addresses these gaps by innovatively combining graph information with tensor representation and providing a general method for uncertainty quantification in tensor-variate classification. While we demonstrate the construction of CP-ROC bands specifically for tensor neural networks, our approach, based on a newly defined similarity metric between tensor representations, can be applied to other classification methods as well.

In this paper, we focus on the classification of graph-structured objects, utilizing GNNs for representation learning alongside topological features, and introduce conditional prediction ROC (CP-ROC) bands for ROC curves that are robust to covariate shifts.

\section{Problem Setups}
An attributed graph $\mathcal{G} = (\mathcal{V}, \mathcal{E}, \boldsymbol{X})$ consists of nodes $\mathcal{V}$, edges $\mathcal{E}$, and node feature matrix $\boldsymbol{X} \in \mathcal{R}^{N \times F}$. Its adjacency matrix $\boldsymbol{A} \in \mathcal{R}^{N \times N}$ has entries $a_{ij} = \omega_{ij}$ for connected nodes (weight $\omega_{ij}\equiv 1$ for unweighted graphs). The degree matrix $\mathcal{D}$ has $d_{ii} = \sum_j a_{ij}$. In graph classification, we have graph-label pairs $\overline{\mathcal{G}}_i = (\mathcal{G}_i, y_i)$. The task is to predict a graph's label. A trained model $\widehat{f}(\mathcal{G})$ outputs $\hat{y}$, the most probable label from $\{1,\dots,L\}$.
Our goal is to obtain a prediction set $C(\mathcal{G},\alpha)$ with confidence level $\alpha\in(0,1)$ for label $y$. We split the dataset $\{\overline{\mathcal{G}}_1, \overline{\mathcal{G}}_2, \dots, \overline{\mathcal{G}}_\aleph\}$ into four disjoint subsets: training $\overline{\mathcal{G}}_{\text{train}}$, validation $\overline{\mathcal{G}}_{\text{valid}}$, calibration $\overline{\mathcal{G}}_{\text{calib}}$, and test $\overline{\mathcal{G}}_{\text{test}}$. The calibration set is reserved for applying conformal prediction for uncertainty quantification.

\textbf{Graph Classification.} Our method is applicable to any trained model $\widehat{f}(\cdot)$. We demonstrate using Tensorized Graph Neural Networks (TTG-NN) \cite{wen2024tensor}, which excels in whole graph classification. TTG-NN captures different levels of local and global representations in real-world graph data. Topological and graphical features of graphs from multi-filtrations and graph convolutions are aggregated together as a high-order tensor feature whose information are extracted automatically with integrated tensor decompositions. We consider the general Tensorized GNN where the input feature and hidden throughput are all in tensor forms. The input feature may be aggregated from more than two channels besides the topological and graphical channel considered in \cite{wen2024tensor}. As a result, $\mathcal{G}_i$ can be viewed as the input tensor feature in the sequel. Mathematical formulations are detailed in Appendix \ref{sec:tensor}. 

\textbf{ROC Curve and AUC.} The \textit{Receiver Operating Characteristic (ROC)} curve evaluates binary classifier performance by plotting True Positive Rate (TPR) against False Positive Rate (FPR) at various classifying thresholds $\lambda\in[0,1]$:
$$
\text{TPR}(\lambda) = \frac{\text{TP}(\lambda)}{\text{TP}(\lambda) + \text{FN}(\lambda)}, \;
\text{FPR}(\lambda) = \frac{\text{FP}(\lambda)}{\text{FP}(\lambda) + \text{TN}(\lambda)}
$$
where $\text{TP}(\lambda)$ (True Positives) is the number of correctly predicted positive instances; $\text{FN}(\lambda)$ (False Negatives) is the number of positive instances incorrectly predicted as negative; $\text{FP}(\lambda)$ (False Positives) is the number of negative instances incorrectly predicted as positive; and $\text{TN}(\lambda)$ (True Negatives) is the number of correctly predicted negative instances.

\textit{Area Under the ROC Curve (AUC)} is a scalar value that summarizes the overall performance of the classifier. It represents the probability that a randomly chosen positive instance is ranked higher than a randomly chosen negative instance. The AUC can be computed as the integral of the ROC curve: 
$$
\text{AUC} = \int_0^1 \text{TPR}(\lambda)\big(\text{FPR}(\lambda)\big) \, d\big(\text{FPR}(\lambda)\big).
$$ 
AUC ranges from 0 to 1, with 1 indicating perfect classification, 0.5 random guessing, and $<0.5$ worse than random.

The ROC curve visually represents classifier performance across thresholds, showing sensitivity-specificity trade-offs. It's particularly useful when false positives and negatives have different consequences. ROC curves enable classifier comparison, especially for imbalanced datasets, with AUC providing a concise performance summary.

\begin{figure}[h]
\centering
\begin{subfigure}{0.4\textwidth}
    \includegraphics[width=\textwidth, height=0.15\textheight]{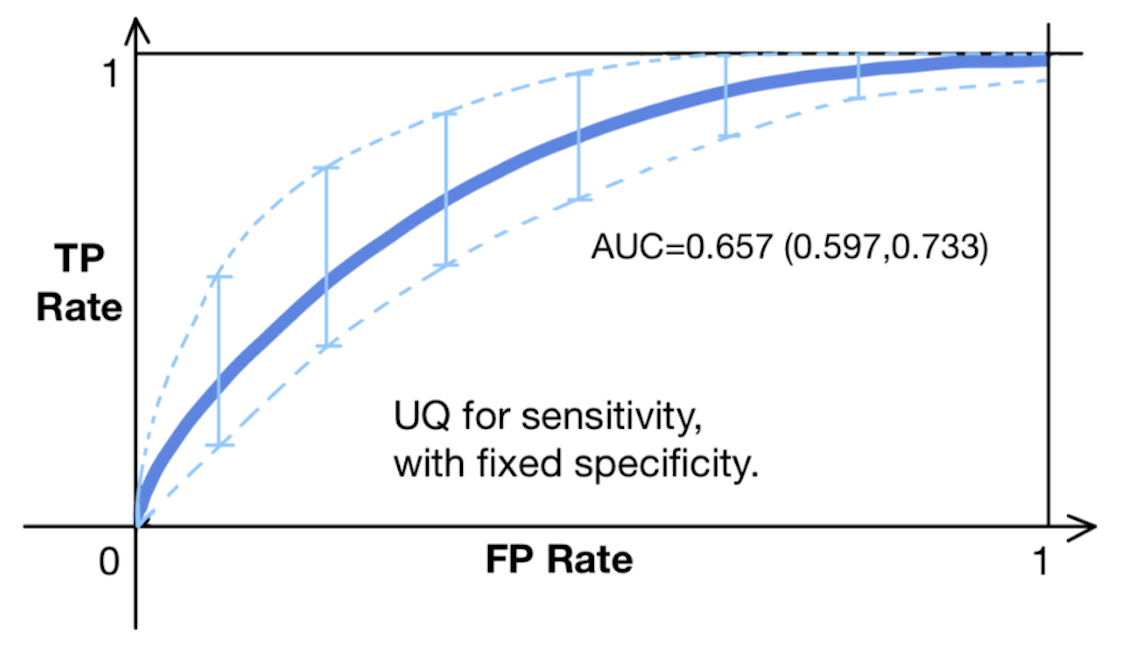}
\end{subfigure}
\begin{subfigure}{0.4\textwidth}
    \includegraphics[width=\textwidth, height=0.15\textheight]{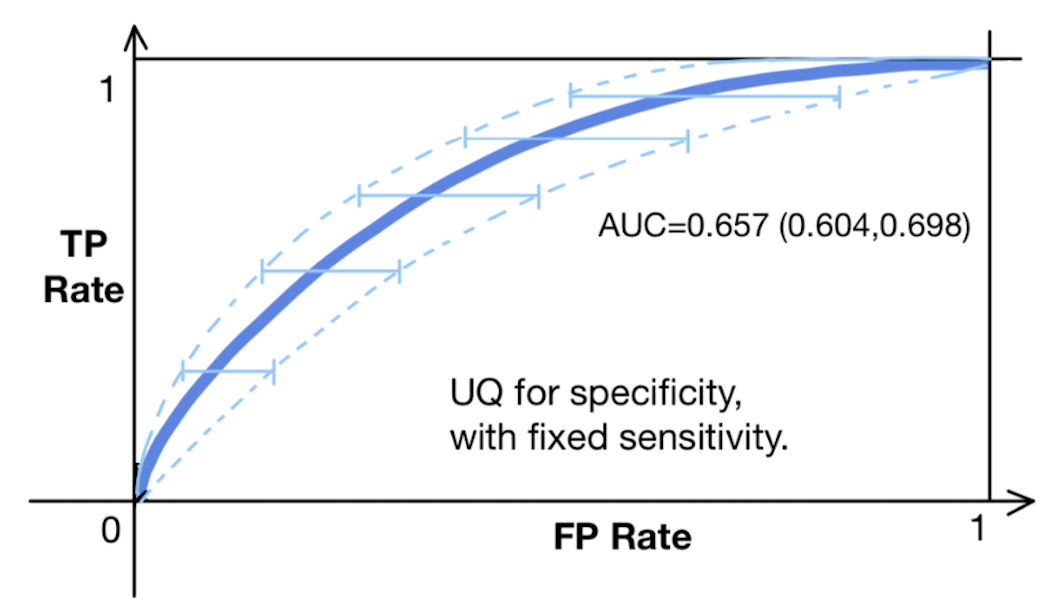}
\end{subfigure}
\caption{\small ROC bands for TPR (Top) and FPR (Bottom).} \label{fig:ROC_bands_example}
\end{figure}

This paper introduces two confidence bands for ROC curve uncertainty quantification: vertical bands for sensitivity (TPR) at fixed FPR, and horizontal bands for specificity (FPR) at fixed TPR, illustrated in Figure \ref{fig:ROC_bands_example}. These bands offer insights into classifier performance for positive and negative groups. AUC's confidence intervals can be derived by calculating the AUC of the upper and lower bounds of these ROC bands. 

\section{Conditional Prediction ROC Bands}
This section introduces our method for constructing ROC bands, first for binary classification and then extending to multi-label classification. In binary classification, graph labels are $0$ or $1$, with an underlying positive probability $\pi(\mathcal{G}) = \mathbb{P} (y=1|\mathcal{G})$. We assume a trained algorithm $\widehat f(\mathcal{G}) = \widehat f(\mathcal{G}|\mathcal{G}_{train})$ is available, estimating $\pi(\mathcal{G})$ based on training data $\mathcal{G}_{train}$.

\subsection{Soft Conformal Prediction}\label{sec: softCP}
Existing conformal prediction methods for predicting labels can result in large confidence set, which is uninformative for binary or few-category cases. To develop ROC bands, we first construct conformal prediction intervals for the soft probability $\pi(\mathcal{G})$, offering a more informative prediction with guaranteed coverage.

Under the common assumption of exchangeability among $\mathcal{G}_{valid}$, $\mathcal{G}_{calib}$, and $\mathcal{G}_{test}$, we construct a prediction interval $C(\mathcal{G})$ for any $\mathcal{G}\in\mathcal{G}_{test}$. Given a user-chosen error rate $\alpha\in[0,1]$, we aim to ensure:
$$
\mathbb{P}(\pi(\mathcal{G}) \in C(\mathcal{G})) \geq 1 - \alpha - o_{n_2}(1)
$$
where $n_2$ is the size of calibration set. 

To construct the prediction interval, we apply a calibration step involving the following procedures: First, we calculate the non-conformity score: $s_i = \tilde\pi(\mathcal{G}_i) - \widehat f(\mathcal{G}_i)$ for $\mathcal{G}_i\in \mathcal{G}_\text{calib}$, where $\tilde\pi(\mathcal{G}_i) $ is a consistent non-parametric estimator of $\pi(\mathcal{G}_i)$. 
Here we use $K$ nearest neighbors for $\tilde\pi(\cdot)$ with distance defined by topological data analysis (TDA) similarity matrix~\cite{TDA}.

Similarly we define $s_{\mathcal{G}}(\pi) = \pi-\hat f(\mathcal{G})$ and a conformal p-value 
$$
p^{\pi}(\mathcal{G}) = \frac{\sum_{j:\mathcal{G}_j\in\mathcal{G}_{calib}} {\bf 1}(s_j<s_{\mathcal{G}}(\pi))+1}{|\mathcal{G}_{calib}|},
$$
indicating a $(1-\alpha)$-level prediction set. Finally, we define the prediction set as 
\[
C(\mathcal{G}) = \{\pi : p^{\pi}(\mathcal{G}) \geq \alpha\}:=
\big[\widehat{f}(\mathcal{G})+q_{\alpha/2}(\{s_i\}),\widehat{f}(\mathcal{G})+q_{1-\alpha/2}(\{s_i\}) \big],
\]
where $s_i$ is the estimated non-conformity score for an object in the calibration set, and $q_{\gamma}(\mathcal{A})$ denotes the $\lfloor\gamma|\mathcal{A}|\rfloor$-th order statistic of set $\mathcal{A}$.

\subsection{CP-ROC Bands for Exchangeable Data}

In this subsection, we leverage the soft conformal prediction interval to construct ROC curve confidence bands under the exchangeable condition. 
Since that each point on the ROC curve represents TPR versus FPR, we need to develop two types of intervals: one for TPR at fixed specificity, and another for FPR at fixed sensitivity. 

For any graph $\mathcal{G}\in\mathcal{G}_{test}$, we modify the previous procedure for developing $C(\mathcal{G})$ to obtain $C_k(\mathcal{G})$, conditioned on label $y=k$. Our aim is to ensure that for any user-chosen error rate $\alpha\in[0,1]$:
$$\mathbb{P}\big(\pi(\mathcal{G}) \in C_k(\mathcal{G})|y=k\big) \geq 1 - \alpha - o_{n}(1),$$
where $n = \min(n_1,n_2)$, with $n_1$ and $n_2$ being the sizes of the training and calibration sets, respectively.

Given $y=k$, $\mathcal{G}$ and $\mathcal{G}_{calib}^k = \{\mathcal{G}_i\in\mathcal{G}_{calib}:y_i=k\}$ are exchangeable, $C_k(\mathcal{G})$ can be obtained from $\big[c_{lo}(\mathcal{G},k),c_{up}(\mathcal{G},k)\big] := $
$$
\big[\widehat{f}(\mathcal{G})+q_{\alpha/2}(\{s_i\}),\widehat{f}(\mathcal{G})+q_{1-\alpha/2}(\{s_i\}_{}) \big],
$$
for $i:\mathcal{G}_i\in\mathcal{G}_{calib}^k$ and $y = k\in\{0,1\}$. 

Next we combine all the confidence intervals together to construct confidence intervals $C_{\lambda}^{sen}(\mathcal{G}_{test}^1)$ for sensitivity ${\rm TPR}(\lambda)$ and $C_{\lambda}^{spe}(\mathcal{G}_{test}^0)$ for specificity ${\rm FPR}(\lambda)$ at any fixed threshold $\lambda\in[0,1]$: 
\begin{align*}
C_{\lambda}^{sen}(\mathcal{G}_{test}^1) &= \bigg[ \frac{1}{|\mathcal{G}_{test}^1|} \sum_{j:\mathcal{G}_j\in\mathcal{G}_{test}^1} \textbf{1}(c_{lo}(\mathcal{G},1)>\lambda),  \frac{1}{|\mathcal{G}_{test}^1|} \sum_{j:\mathcal{G}_j\in\mathcal{G}_{test}^1} \textbf{1}(c_{up}(\mathcal{G},1)>\lambda) \bigg],\\
C_{\lambda}^{spe}(\mathcal{G}_{test}^0) & = \bigg[\frac{1}{|\mathcal{G}_{test}^0|} \sum_{j:\mathcal{G}_j \in \mathcal{G}_{test}^0} \textbf{1}(c_{lo}(\mathcal{G},0) > \lambda), \frac{1}{|\mathcal{G}_{test}^0|} \sum_{j:\mathcal{G}_j \in \mathcal{G}_{test}^0} \textbf{1}(c_{up}(\mathcal{G},0) > \lambda) \bigg].
\end{align*}

Finally, ROC bands can be plotted by collecting 
$C_{\lambda}^{sen}(\mathcal{G}_{test}^1)$ and $C_{\lambda}^{spe}(\mathcal{G}_{test}^0)$ with $\lambda\in[0,1]$.

\subsection{CP-ROC Bands for Non-Exchangeable Data}\label{sec: CP-ROC}

Now we discuss the case where test data are non-exchangeable 
with calibration set, which is our primary goal. In this case, 
we need to modify $C_k(\mathcal{G})$ to the conditional prediction set $C^{cond}_k(\mathcal{G}) $ for any graph $\mathcal{G}\in\mathcal{G}_{test}$ with label $y$, such that for any given a user-chosen error rate $\alpha\in[0,1]$,
$$
\mathbb{P}\big(\pi(\mathcal{G} \in C_k^{cond}(\mathcal{G})\mid \mathcal{G},y=k \big) \geq 1 - \alpha - o_n(1).
$$ 
To guarantee the conditional coverage given $\mathcal{G}_j\in\mathcal{G}_{test}$, our adopted strategy is to replace $\mathcal{G}_{calib}$ by $\mathcal{G}_{calib}^{j}=\{\mathcal{G}_i\in\mathcal{G}_{calib}: \mathcal{G}_i\text{ Similar to }\mathcal{G}_j\}$. While the ``similarity" can be measured in different ways, here we use topology-based similarity matrix to find the $K$ nearest neighbors for our test point. Under the assumption that $\mathcal{G}_j$ is exchangeable with $\mathcal{G}_{calib}^j$, we extend the previous algorithm to the non-exchangeable case by using $\mathcal{G}_{calib}^j$ instead of $\mathcal{G}_{calib}$. 

Specifically, we obtain $C_k^{cond}(\mathcal{G}_j)$: 
$$
\big[c'_{lo}(\mathcal{G}_j,k),c'_{up}(\mathcal{G}_j,k)\big] :=\big[\widehat{f}(\mathcal{G})+q_{\alpha/2}(\{s_i\}),\widehat{f}(\mathcal{G})+q_{1-\alpha/2}(\{s_i\}) \big],
$$
where $i:\mathcal{G}_i\in\mathcal{G}_{calib}^{j,k}$, $\mathcal{G}_{calib}^{j,k} = \{\mathcal{G}_i\in\mathcal{G}_{calib}^{j}:y_i=k\}$ is a subset of $\mathcal{G}_{calib}^{j}$, and $y = k\in\{0,1\}$. Then we can combine all the conditional confidence intervals together to construct confidence intervals $C_{cond,\lambda}^{sen}(\mathcal{G}_{test}^1)$ for sensitivity ${\rm TPR}(\lambda)$ and $C_{cond,\lambda}^{spe}(\mathcal{G}_{test}^0)$ for specificity ${\rm FPR}(\lambda)$ at any fixed threshold $\lambda\in[0,1]$:
\begin{align*}
C_{cond,\lambda}^{sen}(\mathcal{G}_{test}^1) &= \bigg[\frac{1}{|\mathcal{G}_{test}^1|}\sum_{j:\mathcal{G}_j\in\mathcal{G}_{test}^1}{\bf 1}(c'_{lo}(\mathcal{G}_j,1)>\lambda), \frac{1}{|\mathcal{G}_{test}^1|}\sum_{j:\mathcal{G}_j\in\mathcal{G}_{test}^1}{\bf 1}(c'_{up}(\mathcal{G}_j,1)>\lambda)\bigg],\\
C_{cond,\lambda}^{spe}(\mathcal{G}_{test}^0)& = \bigg[\frac{1}{|\mathcal{G}_{test}^0|}\sum_{j:\mathcal{G}_j\in\mathcal{G}_{test}^0}{\bf 1}(c'_{lo}(\mathcal{G}_j,0)>\lambda),\frac{1}{|\mathcal{G}_{test}^0|}\sum_{j:\mathcal{G}_j\in\mathcal{G}_{test}^0}{\bf 1}(c'_{up}(\mathcal{G}_j,0)>\lambda)\bigg].
\end{align*}
Finally, CP-ROC bands can be plotted by collecting 
$C_{cond,\lambda}^{sen}(\mathcal{G}_{test}^1)$ and $C_{cond,\lambda}^{spe}(\mathcal{G}_{test}^0)$, $\lambda\in[0,1]$.

\subsection{Extension to Multi-label Classification}
In this subsection, we consider a multi-label classification model with label $y\in\{1,2,\cdots,L\}$ for the graph $\mathcal{G}$. To plot the ROC curve in this case, the common practice in the literature is to construct separate ROC curves for each label. For a given label $k\in\{1,2,\cdots,L\}$, we define the binary outcome $y_{k,i} = {\bf 1}(y_i=k)$ for any $\mathcal{G}_i$ in the observed data set, and let $\widehat{f}_k(\mathcal{G})$ be the estimated probability that the graph $\mathcal{G}$ is assigned label $k$ by the algorithm trained on the training set. The procedure described in Section~\ref{sec: CP-ROC} can then be applied to visualize the CP-ROC for label $k$ of the multi-label classification model.

\section{Conditional CP-ROC Bands for Tensorized Graph Neural Networks}

We provided a brief overview of the TGNN training process before delving into the construction of CP-ROC bands, which is detailed in Algorithm~\ref{alg:roc}. 

The TGNN, as outlined in Algorithm~\ref{alg:pre-train} in the Appendix, begins with randomly initialized parameters. The dataset is then divided into training and testing\_calibration sets. The model undergoes a typical deep learning training regimen, which includes multiple epochs of batch processing, generating predictions, evaluating Binary Cross-Entropy loss, and updating parameters through backpropagation. This training process is designed to enhance TGNN's predictive accuracy for graph-structured inputs.

The ROC conformal prediction process employs an iterative approach, utilizing multiple splits of the test and calibration pools with varying random seeds. For exchangeable data, the entire calibration set is used for each test graph. For non-exchangeable data, a predefined topology similarity matrix $D$ identifies the $K$ nearest data points to form a new calibration set. Subsequently, for each calibration graph $\mathcal{G}_i$, the similarity matrix $D$ is used again to calculate the mean probability of the $K$ nearest datapoints, enabling the calculation of the non-parametric estimator $\tilde\pi(\mathcal{G}_i)$ for the non-conformity score, as outlined in Section~\ref{sec: softCP}.

\begin{algorithm}
\caption{Pseudo-code of ROC Bands for Tensorized Graph Neural Networks}\label{alg:roc}
\begin{algorithmic}[1] 
\small
\State $\text{train}, \text{test\_calib\_pool} = \text{split(dataset)}$
\State \textbf{Set} learning rate $\alpha$, number of epochs $N$, batch size $B$, number of CP epochs $M$

\Comment{\textcolor{blue}{TGNN Pre-Training}}
\State Call Appendix Algorithm \ref{alg:pre-train} for TGNN pre-training.

\Comment{\textcolor{blue}{ROC conformal prediction step}}
\State \textbf{Load} pre-trained TGNN model $f$, and topology-based similarity distance $D$
\State \textbf{Get} probabilities $p$ for each datapoint in dataset from pre-trained model $f$

\For{each iteration from 1 to $M$}
\State $\text{test, calib} = \text{split(test\_calib\_pool, random\_seed)}$
\For{each test}
\If{not exchangeable}
\State \textbf{Find} $K$ nearest calibration sets using $D$
\State \textbf{Set} $calib$ = $K$ nearest calibration sets
\EndIf
\For{each calib $i$}
\State \textbf{Find} $K$ nearest train sets using $D$

\State $\widehat f(\mathcal{G}_i) = p[i]$
\State $\tilde\pi(\mathcal{G}_i)  = \text{mean}(p[K \text{ nearest train sets}])$
\State $s_i = \tilde\pi(\mathcal{G}_i)-\widehat f(\mathcal{G}_i)$
\EndFor

\State \textbf{Calculate} lower and upper quantiles for sensitivity and specificity adjustments
\EndFor
\EndFor
\end{algorithmic}
\end{algorithm}

It's important to discuss how we calculate the topology similarity distance matrix $D$. In our work, we let $\mathcal{D}_{\mathcal{G}_i}$ and $\mathcal{D}_{\mathcal{G}_j}$ be the persistence diagrams for two graphs $\mathcal{G}_i$ and $\mathcal{G}_j$ by using persistent homology~\cite{wasserman2016topologicaldataanalysis}. The Wasserstein distance between these persistence diagrams, denoted by $\mathcal{W}_p(\mathcal{D}_{\mathcal{G}_i}, \mathcal{D}_{\mathcal{G}_j})$, is calculated as follow:
\begin{equation}
    \mathcal{W}_p(\mathcal{D}_{\mathcal{G}_i}, \mathcal{D}_{\mathcal{G}_j}) = \underset{\gamma \in \Gamma}{\text{inf}} ( \sum_{(x,y) \sim \gamma} \left \| x - y  \right \|_{\infty}^{p})^{1/p},
\end{equation}
Our algorithm's final step involves computing lower and upper quantiles for both sensitivity and specificity. This process establishes robust ROC bands that quantifies the uncertainty in the model’s predictions.

\subsection{Theoretical Gaurantees}\label{CP coverage}
This section focuses on establishing the theoretical coverage of the proposed CP-ROC bands $C_{cond,\lambda}^{sen}(\mathcal{G}_{test}^1) $ and $C_{cond,\lambda}^{spe}(\mathcal{G}_{test}^0)$. Our objective is to quantify the uncertainty of $\widehat f(\mathcal{G})$, where the underlying truth is represented by the oracle probability $\pi(\mathcal{G})$. Consequently, our confidence bands aim to encompass the oracle ${\rm FPR}^{(o)}(\lambda)$ and ${\rm TPR}^{(o)}(\lambda)$, which are defined as follows: 
\begin{align*}
&{\rm TPR}^{(o)}(\lambda) = \frac{\sum_{j:\mathcal{G}_j \in \mathcal{G}_{test}^1} \mathbf{1}(\pi(\mathcal{G}_j) \ge \lambda)}{|\mathcal{G}_{test}^1|},\\
&{\rm FPR}^{(o)}(\lambda) = \frac{\sum_{j: \mathcal{G}_j \in \mathcal{G}_{test}^0} \mathbf{1}(\pi(\mathcal{G}_j) \ge \lambda)}{|\mathcal{D}_{test}^0|}.
\end{align*}
We aim to develop asymptotic $(1-\alpha)$-level coverage for our two target confidence intervals at a fixed confidence level $\alpha$. We also aim to ensure that our developed confidence band consistently encompasses the ROC curve. This requires almost sure coverage of ${\rm FPR}(\lambda)$ and ${\rm TPR}(\lambda)$ based on the TGNN used in practice. We present all our theoretical findings in Theorem~\ref{thm: CP-ROC coverage}, with a detailed proof provided in Appendix \ref{sec:pf_thm}. 

\begin{theorem}\label{thm: CP-ROC coverage}
Assume $\mathcal{G}_{test}$ are i.i.d data set, and each $\mathcal{G}_j\in\mathcal{G}_{test}$ is i.i.d with its $K$-nearest calibration sets $\mathcal{G}_{calib}^j$ and also $K$-nearest training sets $\mathcal{G}_{train}^j$. Let $F_{jk}(\cdot)$ denotes the CDF of $\{s_i\}_{i:\mathcal{G}_i\in\mathcal{N}_j,y_i=k}$, assume $F_{jk}(\cdot)$ is Lipschitz continuous
, then
\begin{align*}
\lim_{|\mathcal{G}_{train}|,|\mathcal{G}_{test}^1|,K\to\infty}P\big({\rm TPR}^{(o)}\big(\lambda\big) \in C_{cond,\lambda}^{sen}(\mathcal{D}_{tst}^1,\alpha)\big)
\ge 1-\alpha,\text{ for $\lambda = \pi(\mathcal{G}_s),\mathcal{G}_s\in\mathcal{G}_{test}^1$}
\end{align*}
\begin{align*}
\lim_{|\mathcal{G}_{train}|,|\mathcal{G}_{test}^0|,K\to\infty}P\big({\rm FPR}^{(o)}\big(\lambda\big) \in C_{cond,\lambda}^{spe}(\mathcal{D}_{tst}^0,\alpha)\big)
\ge 1-\alpha,\text{ for $\lambda = \pi(\mathcal{G}_s),\mathcal{G}_s\in\mathcal{G}_{test}^0$}
\end{align*}
In addition, we assume $F_{jk}^{-1}(\alpha/2)<0<F_{jk}^{-1}(1-\alpha/2) $. Then for any $\lambda\in(0,1)$, as $|\mathcal{G}_{test}^k|,|\mathcal{G}_{train}|,K\to\infty$, almost surely,
\begin{align*}
&{\rm TPR}(\lambda) \in C_{cond,\lambda}^{sen}(\mathcal{D}_{tst}^1,\alpha)\\
&{\rm FPR}(\lambda) \in C_{cond,\lambda}^{spe}(\mathcal{D}_{tst}^0,\alpha)
\end{align*}
\end{theorem}
Notice that by definition, the ROC curve is a step function with jumps at $\lambda = \pi(\mathcal{G})$, $ \mathcal{G} \in \mathcal{G}_{test} $. Thus, the first part of the theorem above shows that if we randomly choose a jump point on the ROC, the confidence interval with fixed sensitivity (for a point with a negative outcome) or with sensitivity (for a point with a positive outcome) will cover the oracle $ {\rm FPR}^{(o)} $ or $ {\rm TPR}^{(o)} $ with confidence level converging to $ 1 - \alpha $. 

As we use the $K$-nearest neighbors to guarantee the exchangeability as well as the consistency of $\tilde\pi(\cdot)$, this algorithm will largely depends on the similarity measure we choose and also the neighbor size $K$. In the theorem, we also requires $K\to\infty$, thus also requiring large size of calibration and training data.

The assumption that CDF $F_{jk}(\cdot)$ is Lipschitz continuous is weak because the underlying probability space of the graph is likely continuous. Finally, for coverage of ${\rm TPR}(\lambda)$ and ${\rm FPR}(\lambda)$, we assume the conformity scores have heavy left tails below $0$ and right tails above $0$. If an algorithm violates this assumption, we can calibrate it according to its prediction error, so that the conformity score is shifted to center at $0$.

\section{Experiments}
\subsection{Experiment Settings}\label{settings}
\noindent{\bf Datasets and Baselines.} We assess the uncertainty quantification performance of the TGNN model on graph classification tasks on chemical compounds and protein molecules. For chemical compounds, the datasets include DHFR, BZR, and COX2~\cite{sutherland2003spline,kriege2012subgraph}, which are composed of graphs representing chemical compounds with nodes as atoms and edges as chemical bonds. DHFR is used for enzyme-ligand binding affinity prediction, BZR for bioactivity against the benzodiazepine receptor, and COX2 for predicting enzyme inhibition, crucial in drug design. In the case of protein molecules, we use datasets including PROTEINS, D\&D, and PTC\_MM~\cite{helma2001predictive,dobson2003distinguishing,borgwardt2005protein,kriege2012subgraph}, where each protein is depicted as a graph with nodes representing amino acids and edges denoting interactions like physical bonds or spatial proximity. These datasets aid in tasks such as protein structure classification and distinguishing chemical compounds based on carcinogenicity in male mice (MM). All selected datasets involve binary classification, making them suitable for testing ROC curve-based UQ methods. Our experimental setup includes a split ratio of 0.8/0.2 for the training and testing\_calibration pool, respectively, and a further split of the testing\_calibration pool into testing and calibration subsets with a ratio of 0.5/0.5. Table~\ref{tab:datasets} in the Appendix summarizes the characteristics of these datasets.
\begin{figure}
\centering
\includegraphics[width=0.85\linewidth]{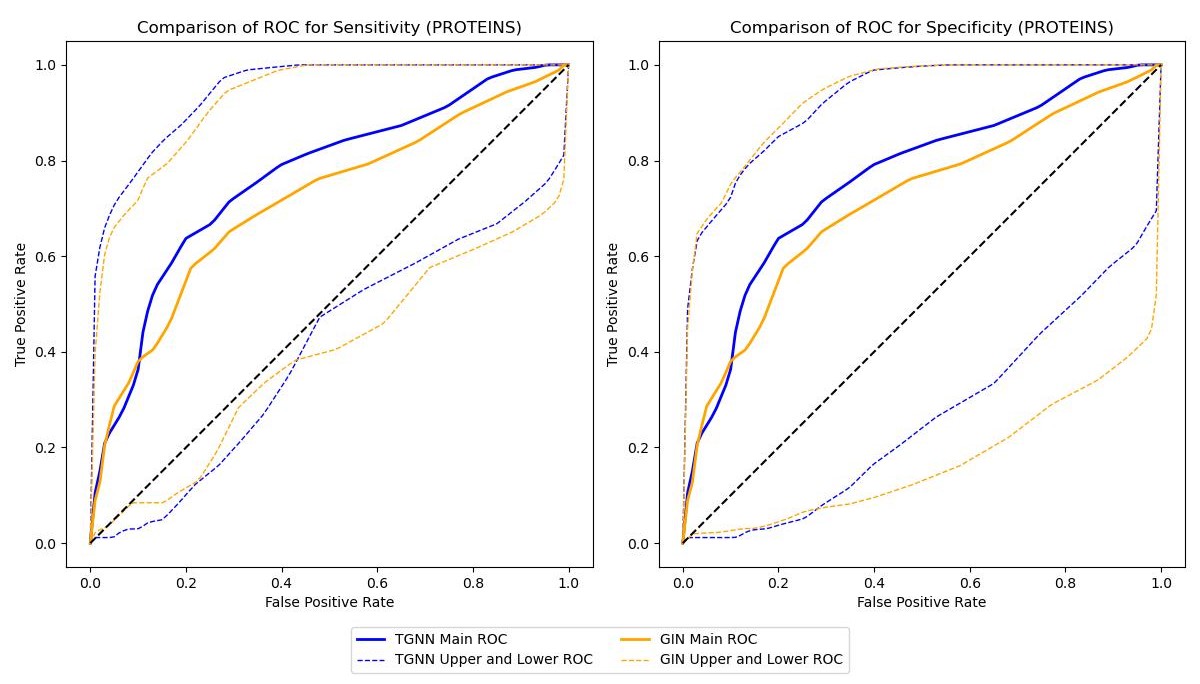}
\caption{\small Example of exchangeable ROC bands for TGNN and GIN.}
\label{fig: ROC_bands}
\end{figure}

\begin{table*}[ht]
    \vspace{-2ex}
    \centering
    \small
    \caption{\small Classification and UQ performance on molecular and chemical graphs. Best results are in \textbf{bold}. \label{classification_result_0_bio_graphs}}
    \resizebox{\textwidth}{!}{
    \begin{tabular}{lcc|cc|cccccc}
    \toprule
    \textbf{{Datasets}}  & \textbf{TGNN\_AUC} & \textbf{{GIN\_AUC}}& \textbf{TGNN\_SEN}(i.i.d, Non i.i.d) & \textbf{{GIN\_SEN}} (i.i.d, Non i.i.d)& \textbf{TGNN\_SPE} (i.i.d, Non i.i.d)& \textbf{{GIN\_SPE}} (i.i.d, Non i.i.d)\\
    \midrule
    \textbf{{BZR}} & \textbf{0.8380}& 0.8267 & \textbf{0.3418}, \textbf{0.3393} & 0.3834, 0.3693 &0.6695, 0.6386 & \textbf{0.5949}, \textbf{0.5611}\\
    \textbf{{COX2}} & \textbf{0.7205}& 0.7054 & \textbf{0.5939}, \textbf{0.4765} & 0.6157, 0.5035 & 0.6712, 0.5712 & \textbf{0.6320}, \textbf{0.5330}\\
    \textbf{{DHFR}} & 0.8514& \textbf{0.8525} & 0.4417, \textbf{0.3573} & \textbf{0.4360}, 0.3645 &\textbf{0.5448}, \textbf{0.4810} & 0.6079, 0.5422\\
    \textbf{{PTC\_MM}} & 0.7191 &\textbf{0.7202} & \textbf{0.6021}, \textbf{0.5223} & 0.6109, 0.5538 & 0.7458, \textbf{0.6186} & \textbf{0.7414}, 0.6249\\
    \textbf{{D\&D}} & \textbf{0.8020}&0.7888 & \textbf{0.4291}, \textbf{0.3724} & 0.4307, 0.3783 & 0.5546, 0.4571& \textbf{0.5326}, \textbf{0.4442}\\
    \textbf{{PROTEINS}} & \textbf{0.7631}&0.7132 & 0.5396, 0.5091&\textbf{0.5378}, \textbf{0.5007} & \textbf{0.6611}, \textbf{0.6156} & 0.7613, 0.7315\\
    \bottomrule
    \end{tabular}}
\end{table*}

\noindent{\bf Experimental Setup.}
We implement our TGNN model within a Pytorch framework on a single NVIDIA Quadro RTX 8000 GPU with a 48GB memory capacity. We employ the Adam optimizer with a learning rate of 0.001 and utilize ReLU as the activation function across the model, except for the Softmax activation in the MLP classifier output. The resolution of the persistence image (PI)~\cite{adams2017persistence} (i.e., we vectorize a persistence diagram $\mathcal{D}_{\mathcal{G}_i}$ into the PI$_{\mathcal{D}_{\mathcal{G}_i}}$) is set to $P = 50$. We explore five different filtrations in our experiments: degree-based, betweenness-based, closeness-based, communicability-based, and eigenvector-based, with batch sizes maintained at 16. The graph convolution layers and MLPs are optimized to determine the best number of hidden units, chosen from $\{16, 32, 64, 128, 256\}$, with TTL featuring 32 hidden units. Our TGNN model incorporates three layers in the graph convolution blocks and two layers in the MLPs, with a consistent dropout rate of 0.5. For conformal prediction, $K = \{20,30,50\}$ neighbors are considered in the $K$-nearest neighbor approach, with an error rate $\alpha$ of 0.1, the exact number of neighbors being contingent upon the dataset size. Please refer to code and data via \url{https://github.com/CS-SAIL/UQ-GNN-ROC.git}.

\subsection{Results}\label{result}
In this section, we showcase experimental results of our novel uncertainty quantification method for ROC curves. As shown in Table~\ref{classification_result_0_bio_graphs}, it highlights the areas under the ROC curve (AUC) for TGNN and GIN models~\cite{xu2018powerful}, i.e., in the first two columns. The latter columns display the ROC bandwidths for sensitivity and specificity under both i.i.d and non-i.i.d conditions. We observe that our proposed CP-UQ method not only enhances the predictive reliability of graph neural networks across various binary classification tasks but extends to other machine learning methods, including logistic regression and SVMs, as well as additional deep learning models. Note that, unlike traditional uncertainty quantification methods which do not offer ROC bands, our approach generates ROC bands for both visual and quantitative measures of uncertainty. 

Additionally, our results reveal that the non-i.i.d ROC bandwidths are consistently smaller than those in i.i.d scenarios. Due to the limited size of training and test dataset, the test data may be inconclusive, which makes the uncertainty by soft conformal prediction under the i.i.d setting larger than it should be. However, our proposed soft conformal prediction under the non-i.i.d settings show good capability handling this condition. For each test point $(\mathcal{G}_{i},y_i)\in\mathcal{G}_{test}$, the subgroup of calibration dataset $\mathcal{N}_{i} = \{(\mathcal{G},y)\in\mathcal{D}_{ca}: dist(\mathcal{G},\mathcal{G}_{i})\le d\}$, which includes similar individuals as the prediction target, are chosen as a substitute of calibration set. $(\mathcal{G}_{i},y_i)$ and corresponding $\mathcal{N}_{i}$ will be approximately i.i.d if $\mathcal{N}_{i}$ is correctly selected such that the uncertainty will be measured accurately \cite{zheng2024roc}. As a result, the uncertainty of our algorithm is smaller using the non-i.i.d soft conformal prediction. 

Our findings also indicate that despite high AUC values from the models, some datasets exhibit moderate uncertainty. The quantification of uncertainty in TGNN is limited by dataset size, and as a compromise, the calibration dataset comprises 20\% data. Soft conformal prediction, which ensures coverage, requires sufficient calibration points to achieve narrow and precise uncertainty estimation. The confidence interval narrows as more points from the target distribution are included in the calibration set. Despite these limitations and the potential for future enhancements with larger datasets, the current performance of the TGNN on these datasets is promising.

The estimated AUC does not always inversely correlate with the algorithm’s uncertainty. While the AUC measures overall performance, including sensitivity and specificity, it is notable that the TGNN model achieves higher AUCs and shorter confidence intervals for sensitivity, which improves hit rates. However, there is a slight increase in the ROC bands and uncertainty for specificity. Essentially, the algorithm more reliably identifies true positives but with a marginal loss in certainty for rejecting false positives. Notably, the maximum increase in the confidence interval for specificity is less than 0.06, whereas the decrease in uncertainty for sensitivity is more pronounced. Moreover, expanding the dataset size is expected to enhance the precision of uncertainty quantification, potentially tightening the ROC bands for specificity further.

The ROC comparisons for the PROTEINS data, as illustrated in the Figure~\ref{fig: ROC_bands}, visually corroborate the numerical findings. That is, we observe that TGNN consistently outperforms GIN across most metrics, demonstrating higher true positive rates and maintaining tighter confidence intervals within the ROC bands. These visual representations align well with our quantitative analyses, reinforcing the superiority of TGNN in handling complex graph-based datasets. Please refer to figures in the Appendix for more details. 

\begin{figure*}[htpb!]
\centering
\includegraphics[width=0.9\textwidth]{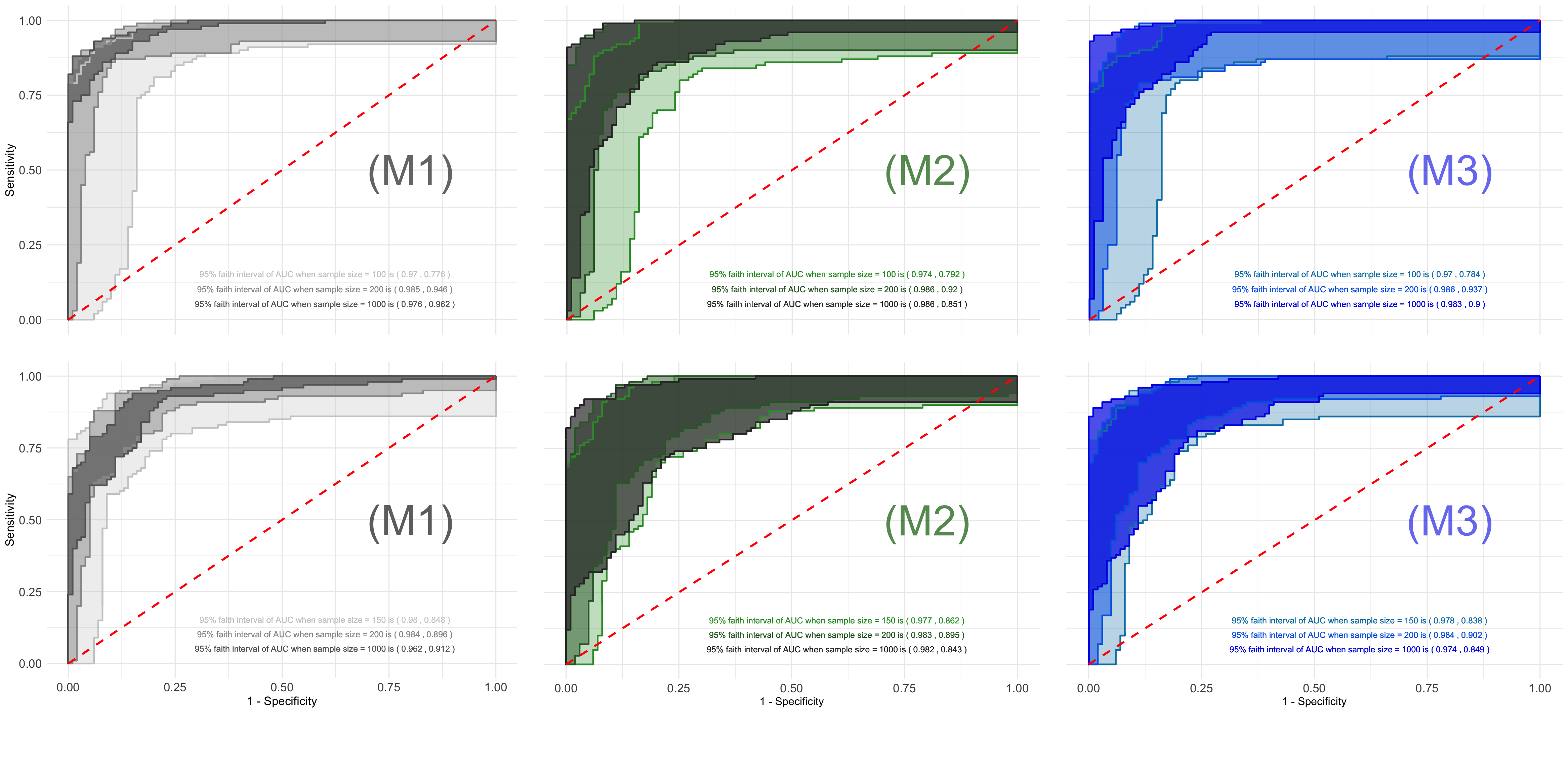}
\caption{Performance comparison among three regression models and varied dataset size. ({\it top}) fixed specificity and ({\it bottom}) fixed sensitivity confidence bands.}
\label{fig: regression}
\end{figure*}
\subsection{Extension to Regression Task}
To further demonstrate the versatility of our approach, we extend its application to regression tasks. This broader scope allows us to validate the robustness of our uncertainty quantification methodology across different types of predictive modeling scenarios beyond graph classification.

For the regression experiments, we simulate datasets by introducing complexities commonly found in real-world applications, such as missing covariates. We investigate three different regression models: a baseline model (M1) with all relevant covariates correctly specified, a model (M2) with one missing covariate, and a model (M3) with two missing covariates. These models are trained on datasets of varying sizes to assess the scalability and stability of our method.

Figure~\ref{fig: regression} shows model performance comparison under both independent (iid) test data using exchangeable CP and non-independent (non-iid) test data using non-exchangeable CP. By examining these different testing conditions, we demonstrate the method’s robustness when applied to both iid and non-iid data, reflecting realistic conditions where data independence cannot always be assumed.

Scalability is a key concern when working with large datasets, and Figure \ref{fig: regression} provides insight into how our proposed uncertainty quantification approach adapts as the training sample size increases. The method shows reliable performance even when covariates are missing in models M2 and M3, maintaining stable uncertainty estimates across varying complexities. Furthermore, to test the robustness of our uncertainty quantification methodology, we applied it to the PROTEINS dataset using a bootstrap method with a resample size of 1000. Figure \ref{fig: bootstrap} in the Appendix presents the 95\% confidence bands of the ROC curves based on bootstrap resampling, alongside the ROC bands generated by our method. We observe that although our area under the ROC curve (AUC) is approximately 0.7 (i.e., moderate predictive power), the uncertainty derived from the bootstrap method is relatively small. This small uncertainty can be misleading, especially given that the dataset we used is imbalanced, which naturally introduces higher uncertainty in model predictability. Additionally, there is no theoretical support for using the bootstrap method in uncertainty quantification, which raises concerns about its validity in these contexts.

By extending our uncertainty quantification methodology to regression tasks, we demonstrate its applicability beyond TGNN-based models. The experimental results, drawn from both synthetic and real-world datasets, indicate that our method can effectively handle missing data and scale efficiently to larger datasets. This extension reinforces the flexibility and robustness of our approach, establishing its potential for a wide range of predictive modeling tasks beyond classification. Moreover, the limitations observed in the bootstrap method further emphasize the importance of our approach, which offers a more theoretically grounded and robust framework for uncertainty quantification.

\section{Conclusion}
This paper introduces CP-ROC bands, a novel method designed for TGNNs to quantify the uncertainty of ROC curves, enhancing reliability in graph classification tasks. CP-ROC, adaptable to various neural networks, addresses challenges in uncertainty quantification due to covariate shifts, and its effectiveness is confirmed through extensive evaluations showing improved prediction reliability and efficiency. In the future, we will extend this idea to the spatio-temporal forecasting tasks.

\clearpage
\bibliography{bib/conformal,bib/ec,bib/neurips_2024,bib/tensor_network,bib/uqgnn}

\begin{thebibliography}{10}

\bibitem{xia2021graph}
Feng Xia, Ke~Sun, Shuo Yu, Abdul Aziz, Liangtian Wan, Shirui Pan, and Huan Liu.
\newblock Graph learning: A survey.
\newblock {\em IEEE Trans AI}, 2(2):109--127, 2021.

\bibitem{zhou2020graph}
Jie Zhou, Ganqu Cui, Shengding Hu, Zhengyan Zhang, Cheng Yang, Zhiyuan Liu, Lifeng Wang, Changcheng Li, and Maosong Sun.
\newblock Graph neural networks: A review of methods and applications.
\newblock {\em AI Open}, 1:57--81, 2020.

\bibitem{wen2024tensor}
Tao Wen, Elynn Chen, and Yuzhou Chen.
\newblock Tensor-view topological graph neural network.
\newblock In {\em International Conference on Artificial Intelligence and Statistics, 2024, Val{\`e}ncia SPAIN}, 2024.

\bibitem{zargarbashi2023conformal}
Soroush~H. Zargarbashi, Simone Antonelli, and Aleksandar Bojchevski.
\newblock Conformal prediction sets for graph neural networks.
\newblock {\em Proceedings of the 40th International Conference on Machine Learning, PMLR 202:12292-12318, 2023.}, 2023.

\bibitem{huang2023uncertainty}
Kexin Huang, Ying Jin, Emmanuel Candès, and Jure Leskovec.
\newblock Uncertainty quantification over graph with conformalized graph neural networks.
\newblock 2023.

\bibitem{zhuang2022uncertainty}
Dingyi Zhuang, Shenhao Wang, Haris Koutsopoulos, and Jinhua Zhao.
\newblock Uncertainty quantification of sparse travel demand prediction with spatial-temporal graph neural networks.
\newblock 2022.

\bibitem{gao2022spatiotemporal}
Xiaowei Gao, Xinke Jiang, Dingyi Zhuang, Huanfa Chen, Shenhao Wang, and James Haworth.
\newblock Spatiotemporal graph neural networks with uncertainty quantification for traffic incident risk prediction.
\newblock 2022.

\bibitem{munikoti2023general}
Sai Munikoti, Deepesh Agarwal, Laya Das, and Balasubramaniam Natarajan.
\newblock A general framework for quantifying aleatoric and epistemic uncertainty in graph neural networks.
\newblock {\em Neurocomputing}, 521:1--10, 2023.

\bibitem{tibshirani2019conformal}
Ryan~J Tibshirani, Rina Foygel~Barber, Emmanuel Candes, and Aaditya Ramdas.
\newblock Conformal prediction under covariate shift.
\newblock {\em Advances in neural information processing systems}, 32, 2019.

\bibitem{gibbs2021adaptive}
Isaac Gibbs and Emmanuel Candes.
\newblock Adaptive conformal inference under distribution shift.
\newblock {\em Advances in Neural Information Processing Systems}, 34:1660--1672, 2021.

\bibitem{vovk2005algorithmic}
Vladimir Vovk, Alexander Gammerman, and Glenn Shafer.
\newblock {\em Algorithmic learning in a random world}, volume~29.
\newblock Springer, 2005.

\bibitem{angelopoulos2020uncertainty}
Anastasios~Nikolas Angelopoulos, Stephen Bates, Michael Jordan, and Jitendra Malik.
\newblock Uncertainty sets for image classifiers using conformal prediction.
\newblock In {\em International Conference on Learning Representations}, 2020.

\bibitem{bates2021distribution}
Stephen Bates, Anastasios Angelopoulos, Lihua Lei, Jitendra Malik, and Michael Jordan.
\newblock Distribution-free, risk-controlling prediction sets.
\newblock {\em Journal of the ACM (JACM)}, 68(6):1--34, 2021.

\bibitem{angelopoulos2022uncertainty}
A.~N.and Stephen~Bates Angelopoulos, Jitendra Malik, and Michael~I. Jordan.
\newblock Uncertainty sets for image classifiers using conformal prediction.
\newblock In {\em Proceedings of the Conference on Neural Information Processing Systems (NeurIPS)}, 2022.

\bibitem{lei2021conformal}
Lihua Lei and Emmanuel~J Cand{\`e}s.
\newblock Conformal inference of counterfactuals and individual treatment effects.
\newblock {\em Journal of the Royal Statistical Society Series B: Statistical Methodology}, 83(5):911--938, 2021.

\bibitem{jin2023sensitivity}
Ying Jin, Zhimei Ren, and Emmanuel~J Cand{\`e}s.
\newblock Sensitivity analysis of individual treatment effects: A robust conformal inference approach.
\newblock {\em Proceedings of the National Academy of Sciences}, 120(6):e2214889120, 2023.

\bibitem{yin2024conformal}
Mingzhang Yin, Claudia Shi, Yixin Wang, and David~M Blei.
\newblock Conformal sensitivity analysis for individual treatment effects.
\newblock {\em Journal of the American Statistical Association}, 119(545):122--135, 2024.

\bibitem{zaffran2022adaptive}
Margaux Zaffran, Olivier F{\'e}ron, Yannig Goude, Julie Josse, and Aymeric Dieuleveut.
\newblock Adaptive conformal predictions for time series.
\newblock In {\em International Conference on Machine Learning}, pages 25834--25866. PMLR, 2022.

\bibitem{jin2023selection}
Ying Jin and Emmanuel~J Cand{\`e}s.
\newblock Selection by prediction with conformal p-values.
\newblock {\em Journal of Machine Learning Research}, 24(244):1--41, 2023.

\bibitem{lei2014distribution}
Jing Lei and Larry Wasserman.
\newblock Distribution-free prediction bands for non-parametric regression.
\newblock {\em Journal of the Royal Statistical Society: Series B}, 76(1):71--96, 2014.

\bibitem{vovk2012conditional}
Vladimir Vovk.
\newblock Conditional validity of inductive conformal predictors.
\newblock In {\em Asian Conference on Machine Learning}, 2012.

\bibitem{barber2021limits}
Rina~Foygel Barber, Emmanuel~J. Cand{\`e}s, Aaditya Ramdas, and Ryan~J. Tibshirani.
\newblock The limits of distribution-free conditional predictive inference.
\newblock {\em Information and Inference}, 10(2):455--482, 2021.

\bibitem{romano2019conformalized}
Yaniv Romano, Evan Patterson, and Emmanuel~J. Cand{\`e}s.
\newblock Conformalized quantile regression.
\newblock In {\em Advances in Neural Information Processing Systems}, 2019.

\bibitem{romano2020classification}
Yaniv Romano, Matteo Sesia, and Emmanuel~J. Cand{\`e}s.
\newblock Classification with valid and adaptive coverage.
\newblock In {\em Advances in Neural Information Processing Systems}, 2020.

\bibitem{angelopoulos2022image}
Anastasios~N. Angelopoulos, Amit~Pal Kohli, Stephen Bates, Michael Jordan, Jitendra Malik, Thayer Alshaabi, Srigokul Upadhyayula, and Yaniv Romano.
\newblock Image-to-image regression with distribution-free uncertainty quantification and applications in imaging.
\newblock In {\em International Conference on Machine Learning}, 2022.

\bibitem{romano2020malice}
Yaniv Romano, Rina~Foygel Barber, Chiara Sabatti, and Emmanuel~J. Cand{\`e}s.
\newblock With malice toward none: {Assessing uncertainty} via equalized coverage.
\newblock {\em Harvard Data Science Review}, 2(2), 2020.

\bibitem{gibbs2023conformal}
Isaac Gibbs, John~J. Cherian, and Emmanuel~J. Cand{\`e}s.
\newblock Conformal prediction with conditional guarantees.
\newblock {\em arXiv preprint arXiv:2305.12616}, 2023.

\bibitem{ding2023class}
T.~Ding, Anastasios~N. Angelopoulos, Stephen Bates, Michael~I. Jordan, and Ryan~J. Tibshirani.
\newblock Class-conditional conformal prediction with many classes.
\newblock In {\em Proceedings of the International Conference on Machine Learning (ICML)}, 2023.

\bibitem{guan2023localized}
Leying Guan.
\newblock Localized conformal prediction: {A} generalized inference framework for conformal prediction.
\newblock {\em Biometrika}, 110(1):33--50, 2023.

\bibitem{schafer1994efficient}
Helmut Schafer.
\newblock Efficient confidence bounds for roc curves.
\newblock {\em Statistics in medicine}, 13(15):1551--1561, 1994.

\bibitem{hilgers1991distribution}
RA~Hilgers.
\newblock Distribution-free confidence bounds for roc curves.
\newblock {\em Methods of information in medicine}, 30(02):96--101, 1991.

\bibitem{jensen2000regional}
Katrin Jensen, Hans-Helge M{\"u}ller, and Helmut Sch{\"a}fer.
\newblock Regional confidence bands for roc curves.
\newblock {\em Statistics in medicine}, 19(4):493--509, 2000.

\bibitem{campbell1994advances}
Gregory Campbell.
\newblock Advances in statistical methodology for the evaluation of diagnostic and laboratory tests.
\newblock {\em Statistics in medicine}, 13(5-7):499--508, 1994.

\bibitem{nakas2023roc}
Christos~T Nakas, Leonidas~E Bantis, and Constantine~A Gatsonis.
\newblock {\em ROC analysis for classification and prediction in practice}.
\newblock Chapman and Hall/CRC, 2023.

\bibitem{adler2009bootstrap}
Werner Adler and Berthold Lausen.
\newblock Bootstrap estimated true and false positive rates and roc curve.
\newblock {\em Computational statistics \& data analysis}, 53(3):718--729, 2009.

\bibitem{zheng2024roc}
Zheshi Zheng, Bo~Yang, and Peter Song.
\newblock Quantifying uncertainty in classification performance: Roc confidence bands using conformal prediction.
\newblock {\em arXiv preprint arXiv:2405.12953}, 2024.

\bibitem{kipf2016semi}
Thomas~N Kipf and Max Welling.
\newblock Semi-supervised classification with graph convolutional networks.
\newblock {\em Proceedings of the International Conference onLearning Representations}, 2017.

\bibitem{velickovic2017graph}
Petar Velickovic, Guillem Cucurull, Arantxa Casanova, Adriana Romero, Pietro Lio, and Yoshua Bengio.
\newblock Graph attention networks.
\newblock {\em arXiv preprint arXiv:1710.10903}, 2017.

\bibitem{gilmer2017neural}
Justin Gilmer, Samuel~S Schoenholz, Patrick~F Riley, Oriol Vinyals, and George~E Dahl.
\newblock Neural message passing for quantum chemistry.
\newblock In {\em Proceedings of the 34th International Conference on Machine Learning}, pages 1263--1272, 2017.

\bibitem{velickovic2018graph}
Petar Veličković and et~al.
\newblock Graph attention networks.
\newblock In {\em International Conference on Learning Representations}, 2018.

\bibitem{li2018deeper}
Qimai Li, Zhichao Han, and Xiao-Ming Wu.
\newblock Deeper insights into graph convolutional networks for semi-supervised learning.
\newblock In {\em Proceedings of the AAAI Conference on Artificial Intelligence}, volume~32, 2018.

\bibitem{zhang2019bayesian}
Ehsan~Hajiramezanali Arman~Hasanzadeh, Shahin Boluki, Mingyuan Zhou, Nick Duffield, Krishna Narayanan, and Xiaoning Qian.
\newblock Bayesian graph neural networks with adaptive connection sampling.
\newblock {\em Proceedings of the 37th International Conference on Machine Learning, Online, PMLR 119, 2020}, 2020.

\bibitem{zhou2013tensor}
Hua Zhou, Lexin Li, and Hongtu Zhu.
\newblock Tensor regression with applications in neuroimaging data analysis.
\newblock {\em Journal of the American Statistical Association}, 108(502):540--552, 2013.

\bibitem{chen2020constrained}
Elynn~Y Chen, Ruey~S Tsay, and Rong Chen.
\newblock Constrained factor models for high-dimensional matrix-variate time series.
\newblock {\em Journal of the American Statistical Association}, 2020.

\bibitem{liu2022identification}
Xialu Liu and Elynn Chen.
\newblock Identification and estimation of threshold matrix-variate factor models.
\newblock {\em Scandinavian Journal of Statistics}, 2022.

\bibitem{chen2022modeling}
Elynn Chen and Rong Chen.
\newblock Modeling dynamic transport network with matrix factor models: with an application to international trade flow.
\newblock {\em Journal of Data Science}, 2022.

\bibitem{bi2018multilayer}
Xuan Bi, Annie Qu, and Xiaotong Shen.
\newblock Multilayer tensor factorization with applications to recommender systems.
\newblock {\em The Annals of Statistics}, 46(6B):3308--3333, 2018.

\bibitem{chen2020modeling}
Elynn Chen, Xin Yun, Rong Chen, and Qiwei Yao.
\newblock Modeling multivariate spatial-temporal data with latent low-dimensional dynamics.
\newblock {\em arXiv preprint arXiv:2002.01305}, 2020.

\bibitem{chen2024semi}
Elynn Chen, Dong Xia, Chencheng Cai, and Jianqing Fan.
\newblock Semi-parametric tensor factor analysis by iteratively projected singular value decomposition.
\newblock {\em Journal of the Royal Statistical Society Series B: Statistical Methodology}, page qkae001, 2024.

\bibitem{chen2024distributed}
Elynn Chen, Xi~Chen, Wenbo Jing, and Yichen Zhang.
\newblock Distributed tensor principal component analysis.
\newblock {\em arXiv preprint arXiv:2405.11681}, 2024.

\bibitem{hao2013linear}
Zhifeng Hao, Lifang He, Bingqian Chen, and Xiaowei Yang.
\newblock A linear support higher-order tensor machine for classification.
\newblock {\em IEEE Transactions on Image Processing}, 22(7):2911--2920, 2013.

\bibitem{guoxian2016}
Xian Guo, Xin Huang, Lefei Zhang, Liangpei Zhang, Antonio Plaza, and Jón~Atli Benediktsson.
\newblock Support tensor machines for classification of hyperspectral remote sensing imagery.
\newblock {\em IEEE Transactions on Geoscience and Remote Sensing}, 54(6):3248--3264, 2016.

\bibitem{chen2024high}
Elynn Chen, Yuefeng Han, and Jiayu Li.
\newblock High-dimensional tensor classification with cp low-rank discriminant structure.
\newblock {\em arXiv preprint arXiv:2409.14397}, 2024.

\bibitem{wimalawarne2016theoretical}
Kishan Wimalawarne, Ryota Tomioka, and Masashi Sugiyama.
\newblock Theoretical and experimental analyses of tensor-based regression and classification.
\newblock {\em Neural Computation}, 28(4):686--715, 2016.

\bibitem{kossaifi2020tensor}
Jean Kossaifi, Zachary~C Lipton, Arinbjorn Kolbeinsson, Aran Khanna, Tommaso Furlanello, and Anima Anandkumar.
\newblock Tensor regression networks.
\newblock {\em Journal of Machine Learning Research}, 21(123):1--21, 2020.

\bibitem{TDA}
Frédéric Chazal and Bertrand Michel.
\newblock An introduction to topological data analysis: fundamental and practical aspects for data scientists, 2021.

\bibitem{wasserman2016topologicaldataanalysis}
Larry Wasserman.
\newblock Topological data analysis, 2016.

\bibitem{sutherland2003spline}
Jeffrey~J Sutherland, Lee~A O'brien, and Donald~F Weaver.
\newblock Spline-fitting with a genetic algorithm: A method for developing classification structure- activity relationships.
\newblock {\em Journal of Chemical Information and Computer Sciences}, 43(6):1906--1915, 2003.

\bibitem{kriege2012subgraph}
Nils Kriege and Petra Mutzel.
\newblock Subgraph matching kernels for attributed graphs.
\newblock In {\em Proceedings of the International Conference on Machine Learning}, pages 291--298, 2012.

\bibitem{helma2001predictive}
Christoph Helma, Ross~D. King, Stefan Kramer, and Ashwin Srinivasan.
\newblock The predictive toxicology challenge 2000--2001.
\newblock {\em Bioinformatics}, 17(1):107--108, 2001.

\bibitem{dobson2003distinguishing}
Paul~D Dobson and Andrew~J Doig.
\newblock Distinguishing enzyme structures from non-enzymes without alignments.
\newblock {\em Journal of Molecular Biology}, 330(4):771--783, 2003.

\bibitem{borgwardt2005protein}
Karsten~M Borgwardt, Cheng~Soon Ong, Stefan Sch{\"o}nauer, SVN Vishwanathan, Alex~J Smola, and Hans-Peter Kriegel.
\newblock Protein function prediction via graph kernels.
\newblock {\em Bioinformatics}, 21:i47--i56, 2005.

\bibitem{adams2017persistence}
Henry Adams, Tegan Emerson, Michael Kirby, Rachel Neville, Chris Peterson, Patrick Shipman, Sofya Chepushtanova, Eric Hanson, Francis Motta, and Lori Ziegelmeier.
\newblock Persistence images: A stable vector representation of persistent homology.
\newblock {\em Journal of Machine Learning Research}, 18, 2017.

\bibitem{xu2018powerful}
Keyulu Xu, Weihua Hu, Jure Leskovec, and Stefanie Jegelka.
\newblock How powerful are graph neural networks?
\newblock In {\em Proceedings of International Conference on Learning Representations}, 2018.

\end{thebibliography}

\clearpage
\begin{center}
{\Large \textbf{Appendix}}
\end{center}

\setcounter{table}{1}
\setcounter{figure}{3}
\setcounter{algorithm}{1}

\section{Proof of Theorem 1} \label{sec:pf_thm}
For the first part of the theorem, we only need to show the coverage rate for $TPR^{(o)}(\lambda)$ and the coverage rate for $FPR^{(o)}(\lambda)$ is exactly the same.
\begin{align}
    &P\big(TPR^{(o)}(\lambda) \in C_{cond,\lambda}^{sen}(\mathcal{G}_{test}^1)\big) = \nonumber\\
    &P\bigg\{
    \frac{1}{|\mathcal{G}_{test}^1|}\sum_{j \in \mathcal{G}_{test}^1} \mathbf{1}\big[c'_{lo}(\mathcal{G}_j,1) > \lambda\big] \leq \frac{1}{|\mathcal{G}_{tst}^1|}\sum_{j \in \mathcal{I}_{tst}^1} \mathbf{1}\big[\pi(\mathcal{G}_j) > \lambda\big]\text{, and }\label{eq:pf3.2_1}  \\
    &\quad \quad\quad
    \frac{1}{|\mathcal{G}_{test}^1|}\sum_{j \in \mathcal{G}_{test}^1} \mathbf{1}\big[\pi(\mathcal{G}_j) > \lambda\big] \leq \frac{1}{|\mathcal{G}_{test}^1|}\sum_{j \in \mathcal{G}_{test}^1} \mathbf{1}\big[c'_{up}(\mathcal{G}_j,1) > \lambda\big]\bigg\} \nonumber
\end{align}
We only need to show the probability bound for one side in Equation~\ref{eq:pf3.2_1}, and the other side will be the same. For the left side,
\begin{align}
    & P\bigg\{ \frac{1}{|\mathcal{G}_{test}^1|}\sum_{j\in\mathcal{G}_{test}^1} {\bf 1}\big[\pi(\mathcal{G}_{j})>\lambda\big] < \frac{1}{|\mathcal{G}_{test}^1|}\sum_{j\in\mathcal{G}_{test}^1} {\bf 1}\big[c'_{lo}(\mathcal{G}_j,1)>\lambda\big]\bigg\}\nonumber\\
    & = P\bigg\{ \sum_{j\in\mathcal{G}_{test}^1} {\bf 1}\big[\pi(\mathcal{G}_{j})>\lambda\big] < \sum_{j\in\mathcal{G}_{test}^1} {\bf 1}\big[ \widehat f(\mathcal{G}_j)+q_{\alpha/2}(\{s_i\}_{i:\mathcal{G}_i\in\mathcal{G}_{calib}^{j,1}}) >\lambda\big]\bigg\}\nonumber
\end{align}
Define $s_j' = \pi(\mathcal{G}_j)-\widehat f(\mathcal{G}_j)$ for $\mathcal{G}_j\in\mathcal{G}_{test}^1$. For the right hand side inside the last term, 
\begin{align*}
   &\sum_{j\in\mathcal{G}_{test}^1} {\bf 1}\big[ \widehat f(\mathcal{G}_j)+q_{\alpha/2}(\{s_i\}_{i:\mathcal{G}_i\in\mathcal{G}_{calib}^{j,k}}) >\lambda\big]\\
   & = \sum_{j\in\mathcal{G}_{test}^1}{\bf 1}\big[ \pi(\mathcal{G}_j)   >\lambda+s_j'- q_{\alpha/2}(\{s_i\}_{i\in\mathcal{G}_{calib}^{j,1}}) \big]{\bf 1}\big[\pi(\mathcal{G}_j) \le \lambda\big]\\
   &\quad\quad\quad\quad+ \big[ \pi(\mathcal{G}_j)   >\lambda+s_j'- q_{\alpha/2}(\{s_i\}_{i\in\mathcal{G}_{calib}^{j,1}}) \big]{\bf 1}\big[\pi(\mathcal{G}_j) > \lambda\big] \\
   &\le \sum_{j\in\mathcal{G}_{test}^1} {\bf 1}\big[s_j'< q_{\alpha/2}(\{s_i\}_{i\in\mathcal{G}_{calib}^{j,1}}) \big]{\bf 1}\big[\pi(\mathcal{G}_j) \le \lambda\big] \\
   &\quad\quad\quad+ {\bf 1}\big[ \pi(\mathcal{G}_j)   >\lambda+s_j'- q_{\alpha/2}(\{ s_i\}_{i\in\mathcal{G}_{calib}^{j,1}}) \big] {\bf 1}\big[\pi(\mathcal{G}_j) > \lambda\big]\\
   &\le \sum_{j\in\mathcal{I}_{tst}^1}{\bf 1}\big[s'_j< q_{\alpha/2}\big(\{s_i\}_{i\in\mathcal{G}_{calib}^{j,1}}\big) \big] \\
   &\quad\quad\quad+ \sqrt{\sum_{j\in\mathcal{I}_{tst}^1} {\bf 1}\big[ \pi(\mathcal{G}_j)   >\lambda+s_j'- q_{\alpha/2}(\{s_i\}_{i\in\mathcal{G}_{calib}^{j,1}}) \big] \sum_{j\in\mathcal{G}_{test}^1}{\bf 1}\big[\pi(\mathcal{G}_j) > \lambda\big]}
\end{align*}
The last inequality comes from Cauchy inequality and also the fact that ${\bf 1}^2(A) = {\bf 1}(A)$ for any event $A$. From the assumption that $\mathcal{G}_j$ are iid with its corresponding $\mathcal{G}_{calib}^j$, so are  $\mathcal{G}_j\in\mathcal{G}_{test}^1$ and $\mathcal{G}_{calib}^1$. 
From the proof of Theorem 3.1 in \cite{zheng2024roc}, we have
\begin{align}
    &\lim_{|\mathcal{G}_{test}^1|,|\mathcal{G}_{train}|,K\to\infty}\frac{1}{|\mathcal{G}_{test}^1|}\sum_{j\in\mathcal{G}_{test}^1}{\bf 1}\big[s'_j< q_{\alpha/2}\big(\{s_i\}_{i\in\mathcal{G}_{calib}^{j,1}}\big) \big] \nonumber\\
    &=\lim_{|\mathcal{G}_{train}|,K\to\infty} \mathbb{E}{\bf 1}(s_j'<q_{\alpha/2}\big(\{s_i\}_{i\in\mathcal{G}_{calib}^{j,1}})<\alpha/2\nonumber
\end{align}

Now plug in $\lambda = \pi(\mathcal{G}_s)$ and we let $|\mathcal{G}_{test}^1|,|\mathcal{G}_{train}|,K\to\infty$, then
\begin{align}
    &P\bigg\{ \frac{1}{|\mathcal{G}_{test}^1|}\sum_{j\in\mathcal{G}_{test}^1} {\bf 1}\big[\pi(\mathcal{G}_{j})>\pi(\mathcal{G}_s)\big] < \frac{1}{|\mathcal{G}_{test}^1|}\sum_{j\in\mathcal{G}_{test}^1} {\bf 1}\big[c'_{lo}(\mathcal{G}_j,1)>\pi(\mathcal{G}_s)\big]\bigg\}\nonumber\\
    &\le P\bigg\{ \frac{1}{|\mathcal{G}_{test}^1|}\sum_{j\in\mathcal{G}_{test}^1} {\bf 1}\big[\pi(\mathcal{G}_{j})>\pi(\mathcal{G}_s)\big]< \alpha/2 + \nonumber\\
    &\quad\quad\quad\quad\sqrt{\sum_{j\in\mathcal{I}_{tst}^1} {\bf 1}\big[ \pi(\mathcal{G}_j)   >\pi(\mathcal{G}_s)+s_j'- q_{\alpha/2}(\{s_i\}_{i\in\mathcal{G}_{calib}^{j,1}}) \big] \sum_{j\in\mathcal{G}_{test}^1}{\bf 1}\big[\pi(\mathcal{G}_j) > \pi(\mathcal{G}_s)\big]}\bigg\}\nonumber\\
   &= \mathbb{E}\frac{1}{|\mathcal{G}_{test}^1|}\sum_{s\in\mathcal{G}_{test}^1}{\bf 1}\bigg\{\frac{1}{|\mathcal{G}_{test}^1|}\sum_{j\in\mathcal{G}_{test}^1} {\bf 1}\big[\pi(\mathcal{G}_{j})>\pi(\mathcal{G}_s)\big]< \alpha/2 + \nonumber\\
    &\quad\quad\quad\quad\sqrt{\sum_{j\in\mathcal{G}_{test}^1} {\bf 1}\big[ \pi(\mathcal{G}_j)   >\pi(\mathcal{G}_s)+s_j'- q_{\alpha/2}(\{s_i\}_{i\in\mathcal{G}_{calib}^{j,1}}) \big] \sum_{j\in\mathcal{G}_{test}^1}{\bf 1}\big[\pi(\mathcal{G}_j) > \pi(\mathcal{G}_s)\big]}\bigg\} \nonumber\\
   &\le \frac{1}{|\mathcal{G}_{test}^1|}\mathbb{E}\sum_{s:q_s\le\alpha/2} 1  = \alpha/2\label{eq:pf3.2_2}
\end{align}
Here $q_s$ denotes the quantile of $\pi(\mathcal{G}_s)$ among $\{\pi(\mathcal{G}_i)\}_{i\in\mathcal{G}_{test}^1}$, and (\ref{eq:pf3.2_2}) comes from the fact that quantile functions are convex. Thus prove the first part of Theorem 3.2. 

For the second part, we only need to prove one side of the inequality in the following:
\begin{align*}
    &\lim_{|\mathcal{G}_{test}^1|,|\mathcal{G}_{train}|,K\to\infty} TPR(\lambda)-\frac{1}{|\mathcal{G}_{test}^1|}\sum_{j \in \mathcal{G}_{test}^1} \mathbf{1}\big[c'_{up}(\mathcal{G}_j,1) > \lambda\big]\\
    &= \lim_{|\mathcal{G}_{test}^1|,|\mathcal{G}_{train}|,K\to\infty} \frac{1}{|\mathcal{G}_{test}^1|} \sum_{j\in\mathcal{G}_{test}^1} \bigg({\bf 1}\big(\widehat f(\mathcal{G}_j)>\lambda\big) - {\bf 1}\big(c'_{up}(\mathcal{G}_j,1)>\lambda\big)\bigg)\\
    &\le \lim_{|\mathcal{G}_{test}^1|,|\mathcal{G}_{train}|,K\to\infty}\frac{1}{|\mathcal{G}_{test}^1|}\sum_{j\in\mathcal{G}_{test}^1}{\bf 1}\big(\widehat f(\mathcal{G}_j)>c'_{up}(\mathcal{G}_j,1)\big) = 0
\end{align*}
The last steps comes from the second part of Theorem 3.1 in \cite{zheng2024roc}.
Similarly, we have the other side of the inequality, thus proof the second part in this theorem.

\section{Tensor Low-Rank Structures}\label{sec:tensor}
Consider an $M$-th order tensor $\mathcal{X}$ of dimension $D_1 \times \cdots \times D_M$. 
If $\mathcal{X}$ assumes a (canonical) rank-$R$ \textit{CP low-rank} structure, then it can be expressed as
\begin{equation} \label{eqn:cp}
    \mathcal{X} = \sum_{r=1}^R c_r \, \mathbf{u}_{1r} \circ \mathbf{u}_{2r} \circ \cdots \mathbf{u}_{Mr},
\end{equation}
where $\circ$ denotes the outer product, $\mathbf{u}_{mr} \in \mathbb{R}^{D_m}$ and $\|\mathbf{u}_{mr}\|_2 = 1$ for all mode $m \in [M]$ and latent dimension $r \in [R]$.
Concatenating all $R$ vectors corresponding to a mode $m$, we have
$\mathbf{U}_m = [\mathbf{u}_{m1}, \cdots, \mathbf{u}_{mR}] \in \mathbb{R}^{D_m \times R}$ which is referred to as the loading matrix for mode $m \in [M]$. 

If $\mathcal{X}$ assumes a rank-$(R_1, \cdots, R_M)$ \textit{Tucker low-rank} structure, then it writes 
\begin{equation*} 
    \mathcal{X} = \mathcal{C} \times_1 \mathbf{U}_1 \times_2 \cdots \times_M \mathbf{U}_M = \sum_{r_1=1}^{R_1} \cdots \sum_{r_M=1}^{R_M} c_{r_1\cdots r_M} (\mathbf{u}_{1 r_1} \circ \cdots \circ \mathbf{u}_{M r_M}),
\end{equation*}
where $\mathbf{u}_{m r_m}$ are all $D_m$-dimensional vectors, and $c_{r_1\cdots r_M}$ are elements in the $R_1 \times \cdots \times R_D$-dimensional core tensor $\mathcal{C}$. 

\textit{Tensor Train (TT) low-rank} approximates a $D_1 \times \cdots \times D_M$ tensor $\mathcal{X}$ with a chain of products of third order \textit{core tensors} $\mathcal{C}_i$, $i\in [M]$, of dimension $R_{i-1} \times D_i \times R_i$.  
Specifically, each element of tensor $\mathcal{X}$ can be written as 
\begin{equation} \label{eqn:tt}
    x_{i_1, \cdots, i_M} = \mathbf{c}_{1,1,i_1,:}^\top
    \times \mathbf{c}_{2,:,i_2,:} \times \cdots \times \mathbf{c}_{M,:,i_M,:}
    \times \mathbf{c}_{M+1,:,1,1}, 
\end{equation}
where $\mathbf{c}_{m,:,i_m,:}$ is an $R_{m-1} \times R_m$ matrix for $m \in [M] \cup \{M+1\}$. 
The product of those matrices is a matrix of size $R_0 \times R_{M+1}$.
Letting $R_0 = 1$, the first core tensor $\mathcal{C}_1$ is of dimension $1 \times D_1 \times R_1$, which is actually a matrix and whose $i_1$-th slice of the middle dimension (i.e., $\mathbf{c}_{1,1,i_1,:}$) is actually a $R_1$ vector. 
To deal with the "boundary condition" at the end, we augmented the chain with an additional tensor $\mathcal{C}_{M+1}$ with $D_{M+1} = 1$ and $R_{M+1} = 1$ of dimension $R_M \times 1 \times 1$.  
So the last tensor can be treated as a vector of dimension $R_M$.

CP low-rank \eqref{eqn:cp} is a special case where the core tensor $\mathcal{C}$ has the same dimensions over all modes, that is $R_m = R$ for all $m\in[M]$, and is super-diagonal.
TT low-rank is a different kind of low-rank structure and it inherits advantages from both CP and Tucker decomposition.
Specifically, TT decomposition can compress tensors as significantly as CP decomposition, while its calculation is as stable as Tucker decomposition. 

\section{Tensorized Graph Neural Networks} \label{sec:TGNN}
The Tensor Transformation Layer (TTL) preserves the tensor structures of feature $\bm{\mathcal{X}}$ of dimension $D=\prod_{m=1}^M D_m$ and hidden throughput. Let $L$ be any positive integer and $\mathbf{d}=\left[d^{(1)}, \cdots, d^{(L+1)}\right]$ collects the width of all layers. A \emph{deep ReLU Tensor Neural Network} is a function mapping taking the form of
\begin{equation} \label{eqn:TNN}
f(\mathcal{X}) = \mathcal{L}^{(L+1)} \circ \sigma \circ \mathcal{L}^{(L)} \circ \sigma \cdots \circ \mathcal{L}^{(2)} \circ \sigma \circ \mathcal{L}^{(1)}(\mathcal{X}), 
\end{equation}
where $\sigma(\cdot)$ is an element-wise activation function. Affine transformation $\mathcal{L}^{(\ell)}(\cdot)$ and hidden input and output tensor of the $\ell$-th layer, i.e., $\mathcal{H}^{(\ell+1)}$ and $\mathcal{H}^{(\ell)}$ are defined by
\begin{equation} \label{trl_eq}
\begin{aligned}
\mathcal{L}^{(\ell)}(\mathcal{H}^{(\ell)}) \triangleq \langle \mathcal{W}^{(\ell)}, \mathcal{H}^{(\ell)} \rangle + \mathcal{B}^{(\ell)}, \\
\text{and}\quad
\mathcal{H}^{(\ell+1)} \triangleq \sigma(\mathcal{L}^{(\ell)}(\mathcal{H}^{(\ell)}))
\end{aligned}
\end{equation}
where
$\mathcal{H}^{(0)} = \bm{\mathcal{X}}$ takes the tensor feature, 
$\langle \cdot, \cdot \rangle$ is the tensor inner product, 
and \emph{low-rank weight} tensor $\bm{\mathcal{W}}^{(\ell)}$ and a bias tensor $\mathcal{B}^{(\ell)}$. 
The tensor structure kicks in when we incorporate tensor low-rank structures such as \textit{CP low-rank}, \textit{Tucker low-rank}, and \textit{Tensor Train low-rank}.

Tucker low-rank structure is defined by
\begin{equation} \label{eqn:tucker}
\mathcal{X} = \mathcal{C} \times_1 \mathbf{U}_1 \times_2 \cdots \times_M \mathbf{U}_M + \mathcal{E},
\end{equation}
where 
$\mathcal{E} \in \mathbb{R}^{D_1 \times \cdots \times D_M}$ is the tensor of the idiosyncratic component (or noise) and 
$\bm{\mathcal{C}}$ is the latent core tensor representing the true low-rank feature tensors and $\mathbf{U}_m$, $m \in [M]$, are the loading matrices.  

The complete definitions of three low-rank structures are given in Appendix~\ref{sec:tensor}. 
CP low-rank \eqref{eqn:cp} is a special case where the core tensor $\mathcal{C}$ has the same dimensions over all modes, that is $R_m = R$ for all $m \in [M]$, and is super-diagonal. 
TT low-rank is a different kind of low-rank structure, which inherits advantages from both CP and Tucker decomposition.
Specifically, TT decomposition can compress tensors as significantly as CP decomposition, while its calculation is as stable as Tucker decomposition.

\section{TGNN Algorithm}
The Algorithm \ref{alg:pre-train} describes the pseudo-code for the training stage of TGNN.
\begin{algorithm}
\caption{Pseudo-code of Pre-training TGNN}\label{alg:pre-train}
\begin{algorithmic}[1] 
\State $train, test\_calib\_pool = \text{split}(dataset)$
\State \textbf{Set} learning rate $\alpha$, number of epochs $N$, batch size $B$, number of CP epochs $M$

\State \textbf{Initialize} TGNN model $f$ with random parameters $\theta$ 

\For{epoch = 1 to $N$}
    \For{batch in \Call{loader}{train, $B$}}
        \State loss = \Call{BCE\_LOSS}{f(batch.inputs), batch.labels}
        \State $\theta = \theta - \alpha \times \Call{f.backward}{loss}$
    \EndFor
\EndFor
\end{algorithmic}
\end{algorithm}
\section{Statistics of Dataset}
Table \ref{tab:datasets} provides some statistics of the benchmark datasets.
\begin{table}[h]
\small
\caption{Statistics of the benchmark datasets.} \label{tab:datasets}
\begin{center}
\setlength{\tabcolsep}{3pt}  
\begin{tabular}{lcccc}
\toprule
\textbf{Dataset} & \textbf{\# Graphs} & \textbf{Avg. \(|\mathcal{V}|\)} & \textbf{Avg. \(|\mathcal{E}|\)}& \textbf{\# Class}\\
\midrule
BZR & 405 & 35.75 & 38.36 &2 \\
COX2 & 467 & 41.22 & 43.45&2 \\
DHFR & 756 & 42.43 & 44.54&2  \\
PTC\_MR & 344 & 14.29 & 14.69&2 \\
D\&D & 1178 & 284.32 & 715.66&2 \\
PROTEINS & 1113 & 39.06 & 72.82&2\\
\bottomrule
\end{tabular}
\end{center}
\end{table}

\section{Boostrap Experiment Figures}
\begin{figure*}[h]
\centering
\includegraphics[width=0.85\textwidth]{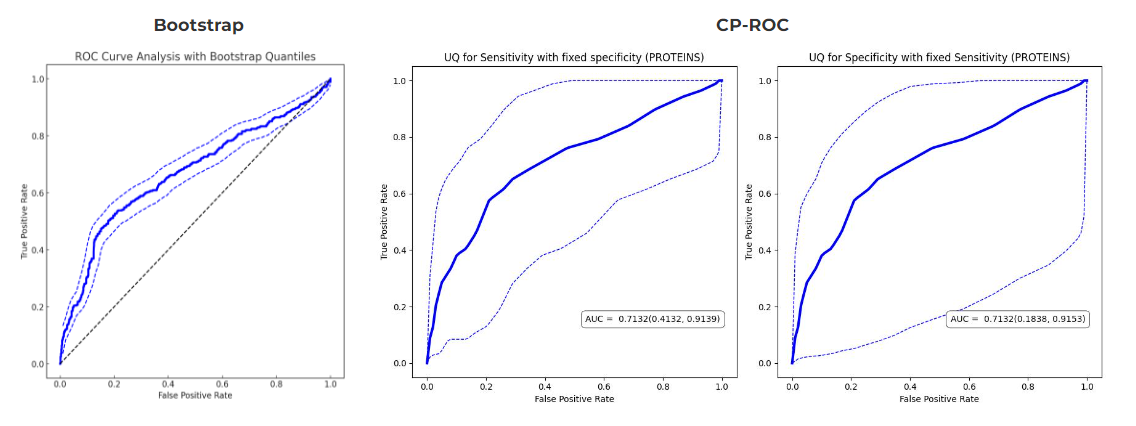}
\caption{\small Comparison between Bootstrap and our UQ on PROTEIN dataset}
\label{fig: bootstrap}
\end{figure*}
Figure \ref{fig: bootstrap} provides a comparison between the traditional Bootstrap method and our proposed uncertainty quantification (UQ) approach on the PROTEIN dataset. The first plot shows the ROC curve with 95\% confidence bands derived from the Bootstrap method, while the second and third plots present the confidence bands generated by our UQ approach using conformal prediction (CP) with fixed specificity and sensitivity, respectively. As demonstrated, the Bootstrap method tends to underestimate uncertainty, particularly in imbalanced datasets, which can mislead model evaluation. Despite the moderate AUC (approximately 0.7), the narrow confidence bands produced by Bootstrap suggest an overly optimistic assessment of the model’s performance. In contrast, our UQ approach provides wider and more realistic confidence bands for poorer-performing models, accurately reflecting the uncertainty in model predictions. This highlights the limitations of the Bootstrap method and the advantages of our UQ framework in offering a more reliable uncertainty estimation for classification tasks.

\section{Supplement ROC comparisons plots} \label{sec:roc_plot}
\begin{figure}[ht]
\centering
\includegraphics[width=1\textwidth]{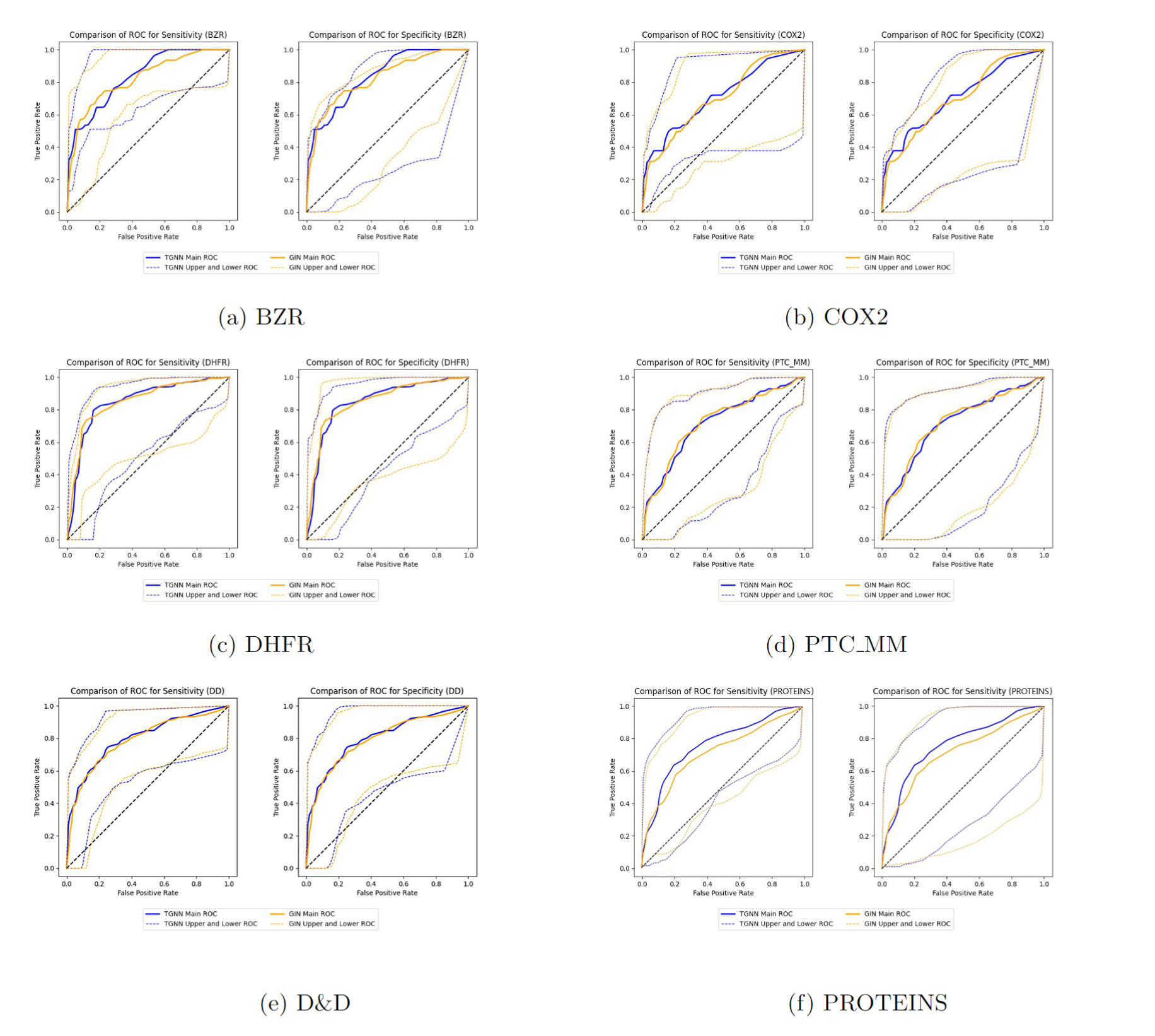}
\caption{\small Exchangeable ROC bands of TGNN and GIN comparison.}
\label{fig:5}
\end{figure}

\begin{figure}[ht!]
\centering
    \begin{subfigure}{0.45\textwidth}
        \centering
        \includegraphics[width=\linewidth]{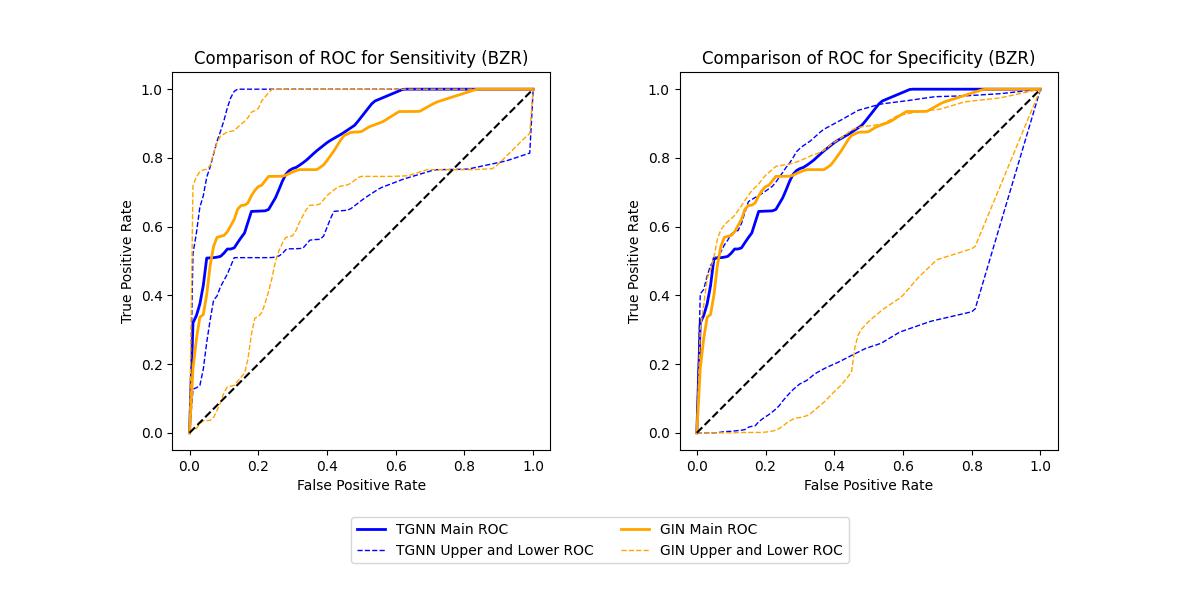}
        \caption{BZR}
    \end{subfigure}
    \begin{subfigure}{0.45\textwidth}
        \centering
        \includegraphics[width=\linewidth]{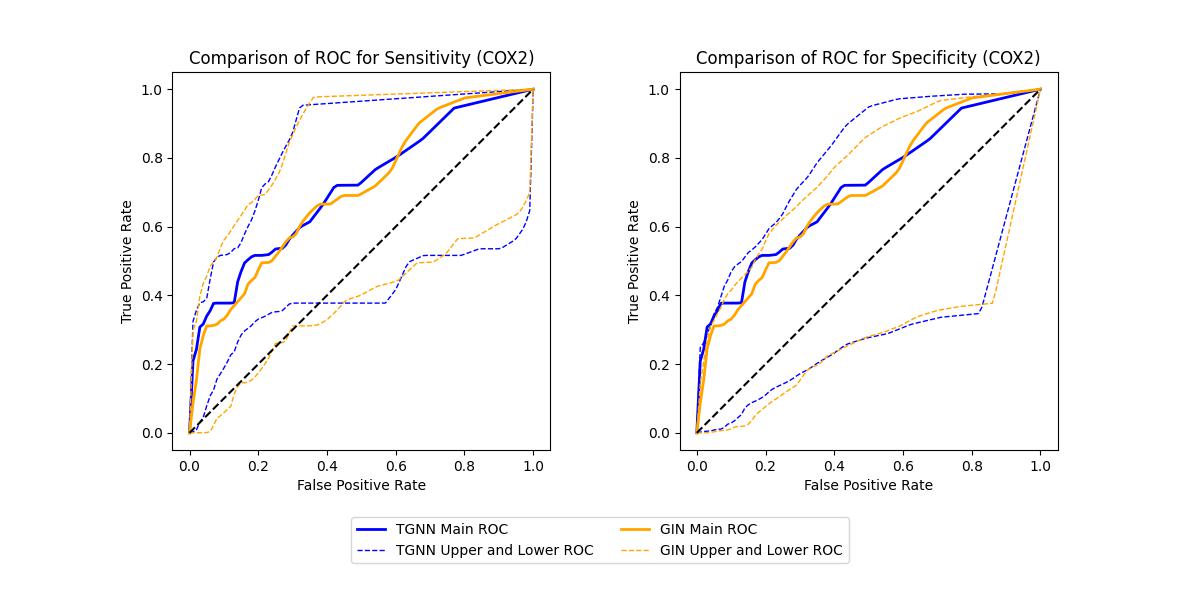}
        \caption{COX2}
    \end{subfigure}

    \begin{subfigure}{0.45\textwidth}
        \centering
        \includegraphics[width=\linewidth]{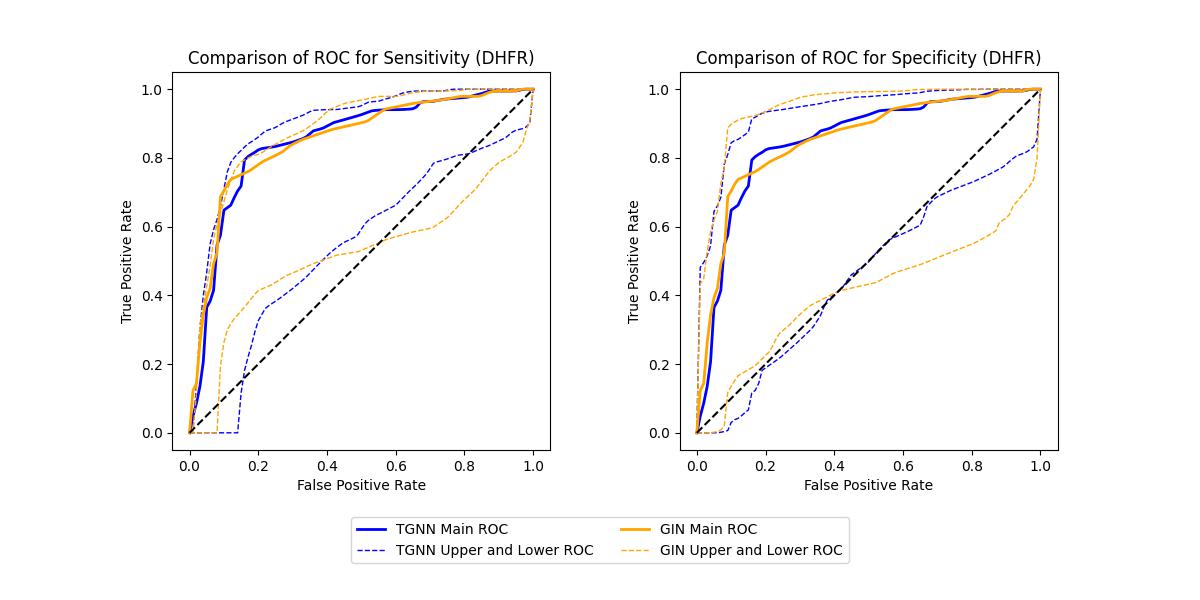}
        \caption{DHFR}
    \end{subfigure}
    \begin{subfigure}{0.45\textwidth}
        \centering
        \includegraphics[width=\linewidth]{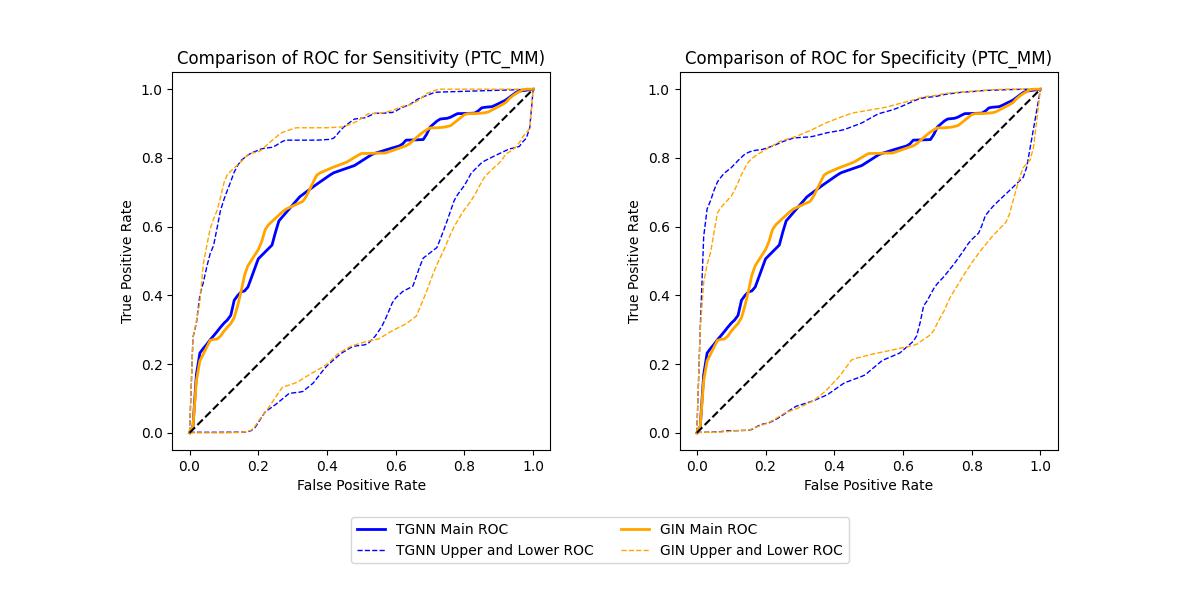}
        \caption{PTC\_MM}
    \end{subfigure}

    \begin{subfigure}{0.45\textwidth}
        \centering
        \includegraphics[width=\linewidth]{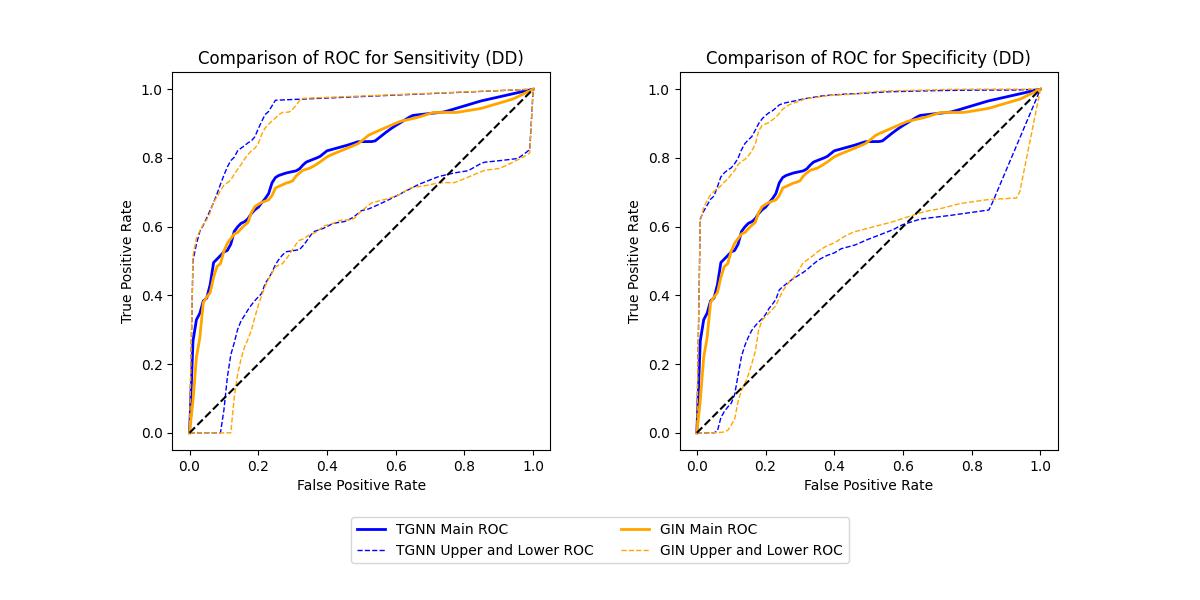}
        \caption{D\&D}
    \end{subfigure}
    \begin{subfigure}{0.45\textwidth}
        \centering
        \includegraphics[width=\linewidth]{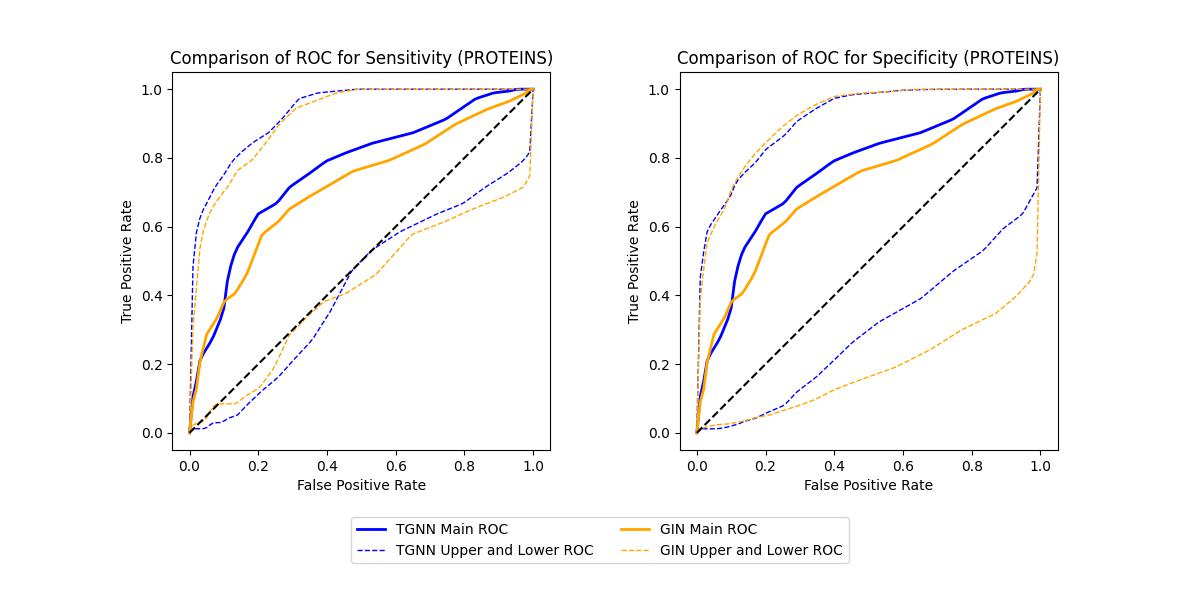}
        \caption{PROTEINS}
    \end{subfigure}
    \caption{\small Non-Exchangeable ROC band for TGNN and GIN comparison.}
\label{fig:6}
\end{figure}

Figures \ref{fig:5} and \ref{fig:6} provide further ROC comparisons across multiple datasets. In most cases, TGNN outperforms GIN, showing higher true positive rates and tighter confidence bands, demonstrating its effectiveness in handling complex graph-based data. However, in some datasets, TGNN exhibits lower performance compared to GIN, which is reflected by larger confidence bands for TGNN. This variation highlights the effectiveness of our UQ methods, accurately capturing model uncertainty and performance fluctuations across different datasets. These figures, together with the quantitative results, emphasize the robustness and versatility of our approach.
\end{document}